\documentclass[10pt, a4paper]{article}

\usepackage[]{lrec-coling2024} 

\usepackage{tikz}
\usepackage{csquotes}

\usepackage[T1]{fontenc}

\usepackage[utf8]{inputenc}

\newcommand\textcyr[1]{{\fontencoding{T2A}\selectfont #1}}
\newcommand\comment[1]{}

\newcommand{\BLU}[1]{{\color{darkblue} {#1}}}

\newcommand{\TRANS}[1]{``{\em {#1}}''}

\newcommand{\STAB}[1]{\begin{tabular}{@{}c@{}}#1\end{tabular}}

\usepackage{microtype}
\usepackage{inconsolata}
\usepackage{gb4e}
\noautomath

\usepackage{multirow}
\usepackage[T2A,OT1]{fontenc}

\newenvironment{itemizerCompact}{\vspace{-1mm}
  \begin{itemize}
    \setlength{\itemsep}{2pt}
    \setlength{\parskip}{0pt}
    \setlength{\parsep}{0pt}

  }
{ \end{itemize}
  \vspace{-1mm}  }

\title{What do Transformers Know about Government?}

\name{Jue Hou,$^{\dagger\star}$
        Anisia Katinskaia,$^{\dagger\star}$
        Lari Kotilainen,$^{\diamondsuit}$\\
        {\bf \large Sathianpong Trangcasanchai,$^{\dagger}$}
        {\bf \large Anh-Duc Vu,$^{\dagger\star}$}
        {\bf \large Roman Yangarber$^{\star}$}
        } 

\address{University of Helsinki, Finland \\
        {\bf \large $^{\dagger}$}Department of Computer Science \\
        {\bf \large $^{\diamondsuit}$}Department of Finnish Language and Culture \\
        {\bf \large $^{\star}$}Department of Digital Humanities\\
         {\tt first.last@helsinki.fi}
         \\}

\abstract{
This paper investigates what insights about linguistic features and what knowledge about the structure of natural language can be obtained from the encodings in transformer language models. \comment{can be used for extending resources.}  
In particular, we explore how BERT encodes the government relation between constituents in a sentence.  We use several probing classifiers, and data from two morphologically rich languages.  Our experiments show that information about government is encoded across all transformer layers, but predominantly in the early layers of the model.  
We find that, for both languages, a small number of attention heads encode enough information about the government relations to enable us to train a classifier capable of discovering new, previously unknown types of government, never seen in the training data.  Currently, data is lacking for the research community working on grammatical constructions, and government in particular.  We release the Government Bank---a dataset defining the government relations for thousands of lemmas in the languages in our experiments.
 \\ \newline  }

\begin{document}

\maketitleabstract

\section{Introduction}

Many contemporary linguistic theories, such as Construction Grammar~\cite{fillmore96,goldberg-2006-construct}, view form-function pairs, called constructions, as fundamental elements of language.  In this paper, our interest in constructions is driven by both theoretical and practical considerations. 
From the theoretical perspective, describing linguistic phenomena in terms of constructions provides a powerful and convenient means of capturing the complex interactions at the interface between lexical semantics and syntax. 

In practical terms, constructions can provide a way of tracking language learners' progress.  Constructions can be seen as describing the linguistic knowledge of a native speaker, and thus as constituting the knowledge needed to master a language---mastery of a language is mastery of its constructions.  If we had a description of the constructions of a language, we could follow a learner's progress in terms of the proportion of constructions s/he has mastered so far; plan learning paths in terms of constructions of increasing complexity, etc.



The present study focuses on one of the most fundamental types of constructions: patterns of government.\footnote{We expect that these methods can be applied to a broader range of constructions, in future work.  We further limit ourselves to {\em verbal} government, in this paper; noun and other types of government to be covered in the future.}
A verb has several syntactic dependents (arguments) in a sentence; they are analyzed as {\em complements} vs.~{\em adjuncts}.
One may say \BLU{``I listened to many songs on a trip through Europe.''}
Here the verb {\em listen} has two prepositional phrase (PP) modifiers: A.~\BLU{``to many songs''}, and B.~\BLU{``on a trip''}. 
Phrase A is a complement---it is semantically {\em required} by the verb; phrase B is an adjunct---it is {\em optional}, in that many actions can take place while ``on a trip,'' nothing binds this PP to this verb in particular.  However, in \BLU{``I went on a trip,''} the same PP is a complement of {\em went}.
Government is about which complements a verb governs: about its {\em valency}---the patterns of complements that a verb expects in a sentence.
Our linguistic competency informs us that one \BLU{\em listens \underline{to}} a sound\,/\,person\,/\,hunch\,/\,etc., while one \BLU{\em goes \underline{on}} a trip\,/\,rampage\,/\,quest\,/\,etc. 
\comment{??? cite COMLEX for English?}

The high performance of pre-trained transformer Language Models (LM) on many NLP tasks has stimulated an interest in investigating the inner representations of these models to find out how linguistic knowledge is encoded inside them, and whether it agrees with linguistic theories in general, and theories of grammar in particular.
Corresponding to our theoretical and practical objectives, this paper explores two Research Questions, respectively---{\bf RQ1}: do transformer LMs encode knowledge about government, and where this knowledge is represented; and {\bf RQ2}: can we {\em extract} this knowledge about government from the LM, to build government resources (or enhance existing resources), e.g., for language learning.
To our knowledge, how LMs encode government has not been explored systematically to date. \comment{I already cited papers about probing different constructions}



To explore RQ1, we probe the BERT model using several probing classifiers, for Finnish and Russian.  We find that information about government relations is encoded across all layers and heads of BERT; however, there is enough information in the initial (5 or 7) layers to detect the presence of the government relations with almost the same accuracy as by using all layers.  Probing classifiers indicate that just a few attention heads encode government relations, and can be used for government prediction without a significant loss in accuracy. 

To explore RQ2, we perform experiments on (a) detecting governing verbs and (b) detecting government patterns, which were held out from training data.  Results show that probing classifiers perform well on both tasks, and therefore can be used to discover {\em new}, previously unseen patterns. 

Descriptions of constructions are important for both general and computational linguistics research, evidenced, e.g., by the ``{\em constructicon}'' efforts~\cite{janda2018constructicon,jbp:/content/journals/10.1075/ijcl.00015.hun,lyngfeld:/content/books/9789027263865}, although the resources resulting from this work are for human consumption, not directly usable, e.g., in CALL systems.\footnote{Computer-aided language learning.} 
Collecting banks of constructions for any language consumes a great deal of labor; we propose a method for searching for patterns in collections of text automatically, by leveraging the information already learned by the LM and encoded in its attention heads.

The paper is organized as follows: 
Section~\ref{sec:prior} briefly reviews prior work on probing neural language models, 
and Section~\ref{sec:background-gov} provides some background on syntactic government.
Section~\ref{sec:data} introduces our government data and its structure.
Section~\ref{sec:experiment} introduces our probing classifier and the experimental setup.
Section~\ref{sec:result} discusses the results of our experiments.
Section~\ref{sec:conclusion} concludes with a discussion of future work.

\section{Related Work}
\label{sec:prior}

Language modeling is a fundamental task in natural language processing (NLP), on which \comment{transformer}pre-trained language models (PLMs) currently achieve the best performance.
One approach to making inferences about the model ``internals'' is through {\em probing}, also known as BERTology~\cite{10.1162/tacl_a_00349}.  Approaches to probing PLMs usually include a specific probing task---e.g., investigating predicate-argument agreement, or how gender is encoded in contextual representations,---data prepared for this task---e.g., minimal pairs of examples that differ only by the studied linguistic category,---and some mechanism that allows us to interact with or query the model's components.  Such a mechanism can be a probing classifier (or \textit{probe}).  
The behavior of a simple\comment{linear or non-linear} probe, trained on representations from the PLM, on a probing task can inform us whether the representations include the linguistic information in question~\cite{adi2017finegrained, conneau-etal-2018-cram, hewitt-manning-2019-structural,10.1609/aaai.v33i01.33016309, hall-maudslay-etal-2020-tale, weissweiler-etal-2022-better, conia-navigli-2022-probing, arps-etal-2022-probing}. 

Some researchers criticize probing classifiers and question their effectiveness, in particular, whether the probed LM in fact uses the information that is discovered by the probe~\cite{hewitt-liang-2019-designing,tamkin-etal-2020-investigating,ravichander-etal-2021-probing,voita-titov-2020-information}.  Later research suggests this criticism can be addressed by designing appropriate control tasks and datasets~\cite{belinkov-2022-probing}.

There are two research directions in probing for dependency relations: token representation or the weights of attention heads.  \citet{doi:10.1073/pnas.1907367117} reconstruct dependency structures based on token representation.  \citet{wu-etal-2020-perturbed} propose parameter-free probing based on masking tokens and measuring the impact of the masked tokens.  They found that a Masked LM (MLM) such as BERT can learn the ``natural'' dependency structure of language.  Although the dependencies learned by the MLM may differ from human annotation or linguistic theory, they consider it a good ``lower bound'' for {\em unsupervised} syntactic parsing.

\citet{doi:10.1073/pnas.1907367117} and \citet{clark-etal-2019-bert} probe attention head weights. They find that no single attention head corresponds well to dependency syntax overall, however, the combination of several attention heads can correspond substantially better than a baseline, where dependency is defined simply by a fixed offset.  Furthermore, they find certain attention heads are {\em specialized} in certain dependency relations.  \citet{kovaleva-etal-2019-revealing} analyze different attention patterns and suggest that some attention heads could potentially be linguistically interpretable.  Our work is inspired by these studies, and we likewise study how information on government is distributed among attention heads. 

\citet{weissweiler-etal-2023-construction} analyze methodologies for probing applied to constructions, as well as probing specifically for certain constructions.  They stress that probing constrictions is challenging for several reasons, in particular, the non-compositional meaning of constructions and the training objectives of PLMs.

\comment{
\begin{table*}[t]
\centering
\scalebox{.9}{
\begin{tabular}{l|l|l|l}
\hline
Lemma & (gloss) & Direct Object & Argument \\
\hline
ajatella		    & \textit{think} & Head:noun + Case:partitive	& Head:noun + Case:elative \\
kyllästyttää	    & \textit{bore} & Head:noun + Case:partitive	& POS:verb + Inf\_Form:inf-A + Case:lative \\
taistella	        & \textit{fight} & --- & HEAD:post-position + Base:puolesta + Case:genitive \\
\hline
\textcyr{готовить}	& \textit{prepare} & Head:noun + Case:accusative &	Head:preposition + Base:\textcyr{к} + Case:dative \\
\textcyr{давать} & \textit{give} & Head:noun + Case:accusative &	Head:noun + Case:dative\\
\hline
\end{tabular}}
\caption{Examples of government relations: verbs with their arguments.  Finnish lemmas ``ajatella'', ``kyllästyttää'', ``taistella''; Russian lemmas ``\textcyr{готовить}'', ``\textcyr{давать}''.  The direct object is displayed as a ``special'' argument, for transitive verbs; other arguments follow. Third and fourth verbs show examples of a post-position ``puolesta'' (\textit{for}) and a preposition ``\textcyr{к}'' (\textit{to}).}
\label{tab:govtable}
\end{table*}
}

\begin{table*}[t]
\centering
\scalebox{1}{
\begin{tabular}{l|l|l|l}
\hline
Lemma & (gloss) & Direct Object & Argument \\
\hline
ajatella		    & \textit{think} & Noun + Case:partitive	& Noun + Case:elative \\
kyllästyttää	    & \textit{bore} & Noun + Case:partitive	& Verb + Inf\_Form:inf-A + Case:lative \\
taistella	        & \textit{fight} & --- & Postposition + Base:puolesta + Case:genitive \\
\hline
\textcyr{готовить}	& \textit{prepare} & Noun + Case:accusative &	Preposition + Base:\textcyr{к} + Case:dative \\
\textcyr{давать} & \textit{give} & Noun + Case:accusative &	Noun + Case:dative\\
\hline
\end{tabular}}
\caption{Examples of government relations: verb lemmas with their complements.
The direct object is displayed as a ``special'' argument, for transitive verbs; other arguments follow. Third and fourth verbs show examples of a post-position ``puolesta'' (\textit{for}) and a preposition ``\textcyr{к}'' (\textit{to}).}
\label{tab:govtable}
\end{table*}

\section{Background on Government}\comment{---Relations}
\label{sec:background-gov}

Government is a relation between a token and its syntactic dependents.  In the context of this paper, the term {\em government} refers to relations between a {\bf governor} (verb, noun, adjective) and a ``\underline{\em governee},'' which can be a noun (phrase), infinitive verb (with its own dependents), an adpositional phrase, etc.  Here are some examples from Finnish:
\vspace{-4px}
\begin{exe}
\ex 
\gll {\bf Ehdotan} \ \underline{teille} pitkää \ \underline{lomaa}.\\
{\em recommend}.V.Pres.1P.Sg \ {\em you}.Pl.Allat {\em long}.Sg.Partit \ {\em break}.Sg.Partit\\
\trans \TRANS{I recommend to you a long break.}
\end{exe}

In example 1, the governor verb ``ehdottaa'' (\textit{recommend}) requires a direct object---a noun phrase ``pitkä loma'' (\textit{long break}), in the partitive case.  The same governor requires another complement---pronoun ``te'' (\textit{you}) to be in the allative case.
\vspace{-4px}
\begin{exe}
\ex
\gll Kuva \ {\bf helpottaa}  \underline{ymmärtämään} \ asiaa. \\
{\em picture}.Sg.Nominat \ {\em ease}.Pres.3P.Sg  {\em understand}.3-Infinit \ {\em thing}.Sg.Partit \\
\trans \TRANS{A picture makes it easier to understand.}
\end{exe}

In example 2, the governor ``helpottaa'' (\textit{ease}) requires its argument---verb ``ymmärtää'' (\textit{to understand}) to be in the illative case of 3-rd infinitive.

\begin{exe}
\ex
\gll Hän  \ {\bf protestoi} päätöstä   \ \underline{vastaan}. \\
{\em S/he}.Sg.Nominat   \ {\em protest}.Past.3P.Sg {\em decision}.Sg.Partit  \ {\em against}.Postpos \\
\trans `S/he protested the decision.'
\end{exe}

In example 3, the governor ``protestoida'' (\textit{protest}) requires the specific postposition ``vastaan'', which in turn governs a noun phrase in partitive case.

In the context of this paper, we focus only on verbs as governors, and use the terms ``complement'', ``argument'', or ``governee'' interchangeably.

\section{Government Data}
\label{sec:data}

As of publish date, we collaborated with expert linguists, and manually collected 1184 Finnish government rules for 765 verb lemmas, and 2635 Russian governments for 1976 verb lemmas,\comment{We will consider imperfective lemmas as unmarked verb forms regarding aspect.} together with several examples for each rule. For each Finnish verb, we include information about its transitivity (whether it is transitive or intransitive) and its government rules, as illustrated in Table~\ref{tab:govtable}. 

Each object or governee indicates the part of speech governed by the verb; note, that the governee can have its own dependents.  \texttt{Case} denotes the case governed by the verb, preposition, or postposition.  Each verb governor can govern more than one argument.   We release this extensive Government Bank, which encodes thousands of government patterns for verbs, nouns, and adjectives---for Finnish and Russian.  The Bank is not ``complete'', but rather encodes the most frequent government patterns in each language.  The patterns are stored in a structured, human-readable form, easily transformed into JSON, tsv, or other formats.\footnote{\href{https://github.com/RevitaAI/GovProbing}{https://github.com/RevitaAI/GovProbing}}

\subsection{Data for Probing Experiments} 

We build a dataset to probe the LM for government relations, using the patterns in the Government Bank.  We start with sentences in the parsed Universal Dependency corpus (v2.12).\comment{Universal Dependency v2.12}  We processed 37K sentences from all Finnish datasets in UD, and 116K Russian sentences from the Taiga~\cite{shavrina2017methodology} and SynTagRus corpora~\cite{droganova2018data}.  We included 27.7K additional Finnish sentences and 102K Russian sentences 
from the WMT News Crawl monolingual training data~\cite{bojar-etal-2017-findings}. 

The data passes through morphological analysis, and parsers---TurkuNLP for Finnish~\cite{udst:turkunlp}, and DeepPavlov for Russian~\cite{burtsev-etal-2018-deeppavlov}.  Then, we filter the sentences and identify valid government instances according to the Government Bank, via rule-based pattern matching.  This results in a dataset with 18K Finnish instances by 582 unique verb lemmas.  For Russian, we collected 143K instances and 2805 unique verbs. 

Using the syntactic parser, we label each dependent of a verb that is also a governee in the Government Bank as a \textbf{positive} instance.  Each sentence can yield one or more positive instances.  As a \textbf{negative} instance, we label any noun phrase or prepositional phrase that is a dependent of the governor\comment{(except the subject) that is {\em not} a complement} but not matching any pattern for this governor's complements in our set of rules. \comment{This is a reasonable assumption because words that are not syntactic dependents of the governor are very unlikely to be its complements.}  This setup helps us avoid the situation where the classifier learns to identify {\em dependency} between two words, which is an easier task than identifying the government relation between the governor verb and its arguments.  We exclude adjunct dependents of the governor from the negative instances since they differ syntactically from complements and may be easier to differentiate. 
We make sure that the positive and negative instances are balanced in the training / test data.

\begin{table}[t]
\begin{tabular}{l|l|cc|cc}
           &     & \multicolumn{2}{c|}{Training} & \multicolumn{2}{c}{Test} \\
  $Dist>3$                & Class  & Far         & Near        & Far         & Near       \\ \hline
\multirow{2}{*}{Finnish} & POS & 402         & 381         & 123         & 144        \\
                         & NEG & 407         & 413         & 118         & 112        \\ \hline
\multirow{2}{*}{Russian} & POS & 3055        & 3047        & 809         & 817        \\
                         & NEG & 3074        & 3125        & 790         & 739       \\
\hline
\end{tabular}
\caption{The number of Finnish and Russian government instances for training and testing when ``far'' means the distance between governor and governee is 3 tokens away.}
\label{tab:num_instances_3}
\end{table}

\begin{table}[t]
\begin{tabular}{l|l|cc|cc}
           &     & \multicolumn{2}{c|}{Training} & \multicolumn{2}{c}{Test} \\
  $Dist>2$                 & Class  & Far         & Near        & Far         & Near       \\ \hline
\multirow{2}{*}{Finnish} & POS & 943         & 925         & 193         & 211        \\
                         & NEG & 999         & 991         & 137         & 145        \\ \hline
\multirow{2}{*}{Russian} & POS & 6905        & 6872        & 2439        & 1746        \\
                         & NEG & 7514        & 7598        & 1830        & 2472       \\
\hline
\end{tabular}
\caption{The number of Finnish and Russian government instances for training and testing when ``far'' means the distance between governor and governee is 2 tokens away.}
\label{tab:num_instances_2}
\end{table}

\subsection{Data Balancing}

The government instances in our datasets are also highly skewed in two respects: in terms of linguistic features (in particular, the case of the governee), and in terms of distance between the governor and governee.  We balance the instances in these respects as well. 

{\bf Linguistic Features:}  
Instances with verbs governing arguments with a certain linguistic feature may outnumber other kinds of instances---the distribution of features of the arguments is not balanced.  For example, the direct object is the most frequent type of argument in Finnish government patterns, which results in a large number of such instances dominating the training data.

{\bf Distance:} We also observe that in most government instances, the governees are adjacent or very near the governor.  To prevent the classifier from learning to classify adjacency rather than government, we divide the instances into two types: ``far'' vs.~``near''.  The governee is ``far'' if its distance to the governor is over $Dist$ complete words (not sub-word tokens); otherwise, we consider it to be ``near''. 
In this paper, we consider two distance thresholds for ``far'' and ``near'': $Dist > 3$ and $Dist > 2$.  The ``near'' instances are much more frequent than ``far''.
We balance the data by down-sampling the ``near'' instances so that the amount of ``near'' and ``far'' instances is about the same. 

{\bf Government Patterns:} To verify whether the classifier is capable of identifying patterns with verbs or arguments that it has never seen before (RQ2), we {\em withhold} instances with certain verb lemmas from the training data, and keep such instances only in the test data.  We also withhold certain argument {\em types}, so that the probing classifier is not trained using all possible types of complements in the Government Bank.  This allows us to evaluate the classifier's ability to discover entirely new, unseen government patterns.  

Table~\ref{tab:num_instances_3} and Table~\ref{tab:num_instances_2} summarize the number of instances used in these experiments, as well as their distance distribution, for Finnish and Russian.\footnote{Detailed data statistics will be in the Appendix~\ref{sec:instance_stat}.}

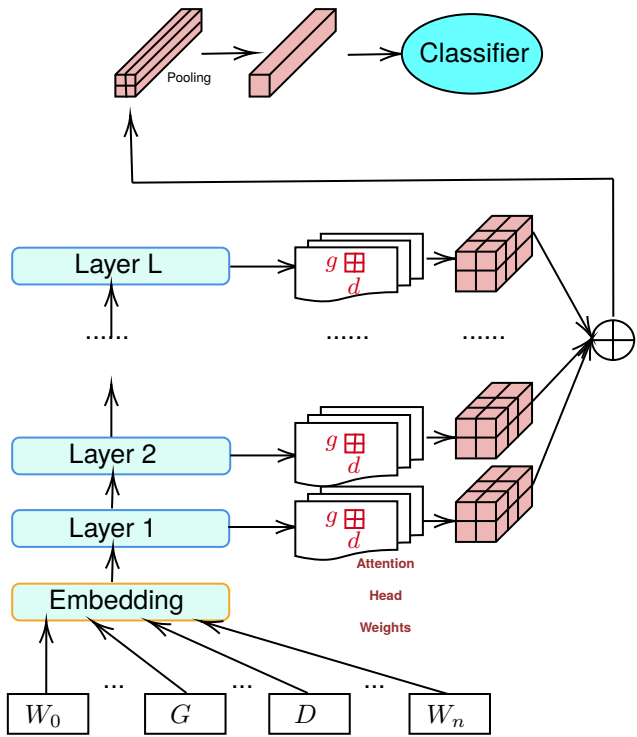
\begin{figure}[t]
\begin{center}

\tikzset{every picture/.style={line width=0.75pt}} 

\begin{tikzpicture}[x=0.75pt,y=0.75pt,yscale=-1,xscale=1]

\draw  [color={rgb, 255:red, 74; green, 144; blue, 226 }  ,draw opacity=1 ][fill={rgb, 255:red, 219; green, 253; blue, 246 }  ,fill opacity=1 ] (35.17,672.4) .. controls (35.17,670.36) and (36.82,668.7) .. (38.87,668.7) -- (139.8,668.7) .. controls (141.84,668.7) and (143.5,670.36) .. (143.5,672.4) -- (143.5,683.5) .. controls (143.5,685.54) and (141.84,687.2) .. (139.8,687.2) -- (38.87,687.2) .. controls (36.82,687.2) and (35.17,685.54) .. (35.17,683.5) -- cycle ;
\draw  [color={rgb, 255:red, 74; green, 144; blue, 226 }  ,draw opacity=1 ][fill={rgb, 255:red, 219; green, 253; blue, 246 }  ,fill opacity=1 ] (35.17,635.9) .. controls (35.17,633.86) and (36.82,632.2) .. (38.87,632.2) -- (140.3,632.2) .. controls (142.34,632.2) and (144,633.86) .. (144,635.9) -- (144,647) .. controls (144,649.04) and (142.34,650.7) .. (140.3,650.7) -- (38.87,650.7) .. controls (36.82,650.7) and (35.17,649.04) .. (35.17,647) -- cycle ;
\draw    (85,667.87) -- (85.44,652.2) ;
\draw [shift={(85.5,650.2)}, rotate = 91.62] [color={rgb, 255:red, 0; green, 0; blue, 0 }  ][line width=0.75]    (10.93,-3.29) .. controls (6.95,-1.4) and (3.31,-0.3) .. (0,0) .. controls (3.31,0.3) and (6.95,1.4) .. (10.93,3.29)   ;
\draw  [color={rgb, 255:red, 74; green, 144; blue, 226 }  ,draw opacity=1 ][fill={rgb, 255:red, 219; green, 253; blue, 246 }  ,fill opacity=1 ] (35.17,540.4) .. controls (35.17,538.36) and (36.82,536.7) .. (38.87,536.7) -- (139.8,536.7) .. controls (141.84,536.7) and (143.5,538.36) .. (143.5,540.4) -- (143.5,551.5) .. controls (143.5,553.54) and (141.84,555.2) .. (139.8,555.2) -- (38.87,555.2) .. controls (36.82,555.2) and (35.17,553.54) .. (35.17,551.5) -- cycle ;
\draw    (84.67,631.87) -- (84.67,607.37) ;
\draw [shift={(84.67,605.37)}, rotate = 90] [color={rgb, 255:red, 0; green, 0; blue, 0 }  ][line width=0.75]    (10.93,-3.29) .. controls (6.95,-1.4) and (3.31,-0.3) .. (0,0) .. controls (3.31,0.3) and (6.95,1.4) .. (10.93,3.29)   ;
\draw    (84.67,581.87) -- (84.67,557.37) ;
\draw [shift={(84.67,555.37)}, rotate = 90] [color={rgb, 255:red, 0; green, 0; blue, 0 }  ][line width=0.75]    (10.93,-3.29) .. controls (6.95,-1.4) and (3.31,-0.3) .. (0,0) .. controls (3.31,0.3) and (6.95,1.4) .. (10.93,3.29)   ;
\draw  [color={rgb, 255:red, 245; green, 166; blue, 35 }  ,draw opacity=1 ][fill={rgb, 255:red, 219; green, 253; blue, 246 }  ,fill opacity=1 ] (35.17,709.4) .. controls (35.17,707.36) and (36.82,705.7) .. (38.87,705.7) -- (139.8,705.7) .. controls (141.84,705.7) and (143.5,707.36) .. (143.5,709.4) -- (143.5,720.5) .. controls (143.5,722.54) and (141.84,724.2) .. (139.8,724.2) -- (38.87,724.2) .. controls (36.82,724.2) and (35.17,722.54) .. (35.17,720.5) -- cycle ;
\draw    (85,704.87) -- (85.44,689.2) ;
\draw [shift={(85.5,687.2)}, rotate = 91.62] [color={rgb, 255:red, 0; green, 0; blue, 0 }  ][line width=0.75]    (10.93,-3.29) .. controls (6.95,-1.4) and (3.31,-0.3) .. (0,0) .. controls (3.31,0.3) and (6.95,1.4) .. (10.93,3.29)   ;
\draw   (33,761.2) -- (73.17,761.2) -- (73.17,781.2) -- (33,781.2) -- cycle ;
\draw   (101.5,761.2) -- (141.67,761.2) -- (141.67,781.2) -- (101.5,781.2) -- cycle ;
\draw   (163.5,761.2) -- (203.67,761.2) -- (203.67,781.2) -- (163.5,781.2) -- cycle ;
\draw   (233.5,761.2) -- (273.67,761.2) -- (273.67,781.2) -- (233.5,781.2) -- cycle ;
\draw    (52.17,761.2) -- (52.2,726.6) ;
\draw [shift={(52.2,724.6)}, rotate = 90.05] [color={rgb, 255:red, 0; green, 0; blue, 0 }  ][line width=0.75]    (10.93,-3.29) .. controls (6.95,-1.4) and (3.31,-0.3) .. (0,0) .. controls (3.31,0.3) and (6.95,1.4) .. (10.93,3.29)   ;
\draw    (122.17,760.7) -- (76.79,725.82) ;
\draw [shift={(75.2,724.6)}, rotate = 37.55] [color={rgb, 255:red, 0; green, 0; blue, 0 }  ][line width=0.75]    (10.93,-3.29) .. controls (6.95,-1.4) and (3.31,-0.3) .. (0,0) .. controls (3.31,0.3) and (6.95,1.4) .. (10.93,3.29)   ;
\draw    (182.67,760.7) -- (105.05,725.4) ;
\draw [shift={(103.23,724.57)}, rotate = 24.46] [color={rgb, 255:red, 0; green, 0; blue, 0 }  ][line width=0.75]    (10.93,-3.29) .. controls (6.95,-1.4) and (3.31,-0.3) .. (0,0) .. controls (3.31,0.3) and (6.95,1.4) .. (10.93,3.29)   ;
\draw    (252.67,761.2) -- (131.15,725.14) ;
\draw [shift={(129.23,724.57)}, rotate = 16.53] [color={rgb, 255:red, 0; green, 0; blue, 0 }  ][line width=0.75]    (10.93,-3.29) .. controls (6.95,-1.4) and (3.31,-0.3) .. (0,0) .. controls (3.31,0.3) and (6.95,1.4) .. (10.93,3.29)   ;
\draw    (143,677.2) -- (175.67,677.2) ;
\draw [shift={(177.67,677.2)}, rotate = 180] [color={rgb, 255:red, 0; green, 0; blue, 0 }  ][line width=0.75]    (10.93,-3.29) .. controls (6.95,-1.4) and (3.31,-0.3) .. (0,0) .. controls (3.31,0.3) and (6.95,1.4) .. (10.93,3.29)   ;
\draw  [fill={rgb, 255:red, 255; green, 255; blue, 255 }  ,fill opacity=1 ] (189.23,656.86) -- (240.17,656.86) -- (240.17,681.28) .. controls (208.33,681.28) and (214.7,690.08) .. (189.23,684.39) -- cycle ; \draw  [fill={rgb, 255:red, 255; green, 255; blue, 255 }  ,fill opacity=1 ] (182.87,660.56) -- (233.8,660.56) -- (233.8,684.98) .. controls (201.97,684.98) and (208.33,693.78) .. (182.87,688.09) -- cycle ; \draw  [fill={rgb, 255:red, 255; green, 255; blue, 255 }  ,fill opacity=1 ] (176.5,664.26) -- (227.43,664.26) -- (227.43,688.68) .. controls (195.6,688.68) and (201.97,697.48) .. (176.5,691.79) -- cycle ;
\draw    (144,641.2) -- (176.67,641.2) ;
\draw [shift={(178.67,641.2)}, rotate = 180] [color={rgb, 255:red, 0; green, 0; blue, 0 }  ][line width=0.75]    (10.93,-3.29) .. controls (6.95,-1.4) and (3.31,-0.3) .. (0,0) .. controls (3.31,0.3) and (6.95,1.4) .. (10.93,3.29)   ;
\draw    (144,546.2) -- (176.67,546.2) ;
\draw [shift={(178.67,546.2)}, rotate = 180] [color={rgb, 255:red, 0; green, 0; blue, 0 }  ][line width=0.75]    (10.93,-3.29) .. controls (6.95,-1.4) and (3.31,-0.3) .. (0,0) .. controls (3.31,0.3) and (6.95,1.4) .. (10.93,3.29)   ;
\draw  [fill={rgb, 255:red, 239; green, 182; blue, 182 }  ,fill opacity=1 ] (256.67,664.7) -- (274.67,646.7) -- (294.67,646.7) -- (294.67,666.7) -- (276.67,684.7) -- (256.67,684.7) -- cycle ; \draw   (294.67,646.7) -- (276.67,664.7) -- (256.67,664.7) ; \draw   (276.67,664.7) -- (276.67,684.7) ;
\draw    (266.67,664.2) -- (266.67,684.2) ;
\draw    (256.67,674.2) -- (276.67,674.2) ;
\draw    (266.67,664.2) -- (284.17,646.7) ;
\draw    (276.67,674.2) -- (295.17,655.7) ;
\draw    (242,542.2) -- (254.67,542.2) ;
\draw [shift={(256.67,542.2)}, rotate = 180] [color={rgb, 255:red, 0; green, 0; blue, 0 }  ][line width=0.75]    (10.93,-3.29) .. controls (6.95,-1.4) and (3.31,-0.3) .. (0,0) .. controls (3.31,0.3) and (6.95,1.4) .. (10.93,3.29)   ;
\draw    (242,632.2) -- (254.67,632.2) ;
\draw [shift={(256.67,632.2)}, rotate = 180] [color={rgb, 255:red, 0; green, 0; blue, 0 }  ][line width=0.75]    (10.93,-3.29) .. controls (6.95,-1.4) and (3.31,-0.3) .. (0,0) .. controls (3.31,0.3) and (6.95,1.4) .. (10.93,3.29)   ;
\draw    (240,674.2) -- (254.67,674.2) ;
\draw [shift={(256.67,674.2)}, rotate = 180] [color={rgb, 255:red, 0; green, 0; blue, 0 }  ][line width=0.75]    (10.93,-3.29) .. controls (6.95,-1.4) and (3.31,-0.3) .. (0,0) .. controls (3.31,0.3) and (6.95,1.4) .. (10.93,3.29)   ;
\draw   (324.5,583.2) .. controls (324.5,577.68) and (329.24,573.2) .. (335.08,573.2) .. controls (340.93,573.2) and (345.67,577.68) .. (345.67,583.2) .. controls (345.67,588.72) and (340.93,593.2) .. (335.08,593.2) .. controls (329.24,593.2) and (324.5,588.72) .. (324.5,583.2) -- cycle ; \draw   (324.5,583.2) -- (345.67,583.2) ; \draw   (335.08,573.2) -- (335.08,593.2) ;
\draw    (295.17,528.7) -- (323.7,580.45) ;
\draw [shift={(324.67,582.2)}, rotate = 241.13] [color={rgb, 255:red, 0; green, 0; blue, 0 }  ][line width=0.75]    (10.93,-3.29) .. controls (6.95,-1.4) and (3.31,-0.3) .. (0,0) .. controls (3.31,0.3) and (6.95,1.4) .. (10.93,3.29)   ;
\draw    (295.17,613.7) -- (323.3,583.66) ;
\draw [shift={(324.67,582.2)}, rotate = 133.12] [color={rgb, 255:red, 0; green, 0; blue, 0 }  ][line width=0.75]    (10.93,-3.29) .. controls (6.95,-1.4) and (3.31,-0.3) .. (0,0) .. controls (3.31,0.3) and (6.95,1.4) .. (10.93,3.29)   ;
\draw    (295.17,655.7) -- (323.92,584.06) ;
\draw [shift={(324.67,582.2)}, rotate = 111.87] [color={rgb, 255:red, 0; green, 0; blue, 0 }  ][line width=0.75]    (10.93,-3.29) .. controls (6.95,-1.4) and (3.31,-0.3) .. (0,0) .. controls (3.31,0.3) and (6.95,1.4) .. (10.93,3.29)   ;
\draw  [fill={rgb, 255:red, 239; green, 182; blue, 182 }  ,fill opacity=1 ] (86.67,449.7) -- (119.67,416.7) -- (129.67,416.7) -- (129.67,427.7) -- (96.67,460.7) -- (86.67,460.7) -- cycle ; \draw   (129.67,416.7) -- (96.67,449.7) -- (86.67,449.7) ; \draw   (96.67,449.7) -- (96.67,460.7) ;
\draw    (91.67,449.2) -- (92.17,461.2) ;
\draw    (86.67,455.2) -- (96.67,455.2) ;
\draw    (91.67,449.2) -- (124.67,416.7) ;
\draw    (96.67,455.2) -- (129.17,423.2) ;
\draw    (335.23,571.43) -- (335.14,502) ;
\draw  [fill={rgb, 255:red, 239; green, 182; blue, 182 }  ,fill opacity=1 ] (153.67,449.7) -- (186.67,416.7) -- (196.67,416.7) -- (196.67,427.7) -- (163.67,460.7) -- (153.67,460.7) -- cycle ; \draw   (196.67,416.7) -- (163.67,449.7) -- (153.67,449.7) ; \draw   (163.67,449.7) -- (163.67,460.7) ;
\draw    (129.67,439.2) -- (154.67,438.74) ;
\draw [shift={(156.67,438.7)}, rotate = 178.94] [color={rgb, 255:red, 0; green, 0; blue, 0 }  ][line width=0.75]    (10.93,-3.29) .. controls (6.95,-1.4) and (3.31,-0.3) .. (0,0) .. controls (3.31,0.3) and (6.95,1.4) .. (10.93,3.29)   ;
\draw    (202.67,439.7) -- (227.67,439.24) ;
\draw [shift={(229.67,439.2)}, rotate = 178.94] [color={rgb, 255:red, 0; green, 0; blue, 0 }  ][line width=0.75]    (10.93,-3.29) .. controls (6.95,-1.4) and (3.31,-0.3) .. (0,0) .. controls (3.31,0.3) and (6.95,1.4) .. (10.93,3.29)   ;
\draw  [fill={rgb, 255:red, 103; green, 255; blue, 255 }  ,fill opacity=1 ] (229.67,439.2) .. controls (229.67,428.15) and (245.34,419.2) .. (264.67,419.2) .. controls (284,419.2) and (299.67,428.15) .. (299.67,439.2) .. controls (299.67,450.25) and (284,459.2) .. (264.67,459.2) .. controls (245.34,459.2) and (229.67,450.25) .. (229.67,439.2) -- cycle ;
\draw    (261.67,659.2) -- (281.67,659.2) ;
\draw    (266.67,654.2) -- (286.67,654.2) ;
\draw    (282.67,659.2) -- (282.67,679.2) ;
\draw    (287.67,654.2) -- (287.67,674.2) ;
\draw  [fill={rgb, 255:red, 239; green, 182; blue, 182 }  ,fill opacity=1 ] (256.67,622.7) -- (274.67,604.7) -- (294.67,604.7) -- (294.67,624.7) -- (276.67,642.7) -- (256.67,642.7) -- cycle ; \draw   (294.67,604.7) -- (276.67,622.7) -- (256.67,622.7) ; \draw   (276.67,622.7) -- (276.67,642.7) ;
\draw    (266.67,622.2) -- (266.67,642.2) ;
\draw    (256.67,632.2) -- (276.67,632.2) ;
\draw    (266.67,622.2) -- (284.17,604.7) ;
\draw    (276.67,632.2) -- (295.17,613.7) ;
\draw    (261.67,617.2) -- (281.67,617.2) ;
\draw    (266.67,612.2) -- (286.67,612.2) ;
\draw    (282.67,617.2) -- (282.67,637.2) ;
\draw    (287.67,612.2) -- (287.67,632.2) ;
\draw  [fill={rgb, 255:red, 239; green, 182; blue, 182 }  ,fill opacity=1 ] (256.67,538.7) -- (274.67,520.7) -- (294.67,520.7) -- (294.67,540.7) -- (276.67,558.7) -- (256.67,558.7) -- cycle ; \draw   (294.67,520.7) -- (276.67,538.7) -- (256.67,538.7) ; \draw   (276.67,538.7) -- (276.67,558.7) ;
\draw    (266.67,538.2) -- (266.67,558.2) ;
\draw    (256.67,548.2) -- (276.67,548.2) ;
\draw    (266.67,538.2) -- (284.17,520.7) ;
\draw    (276.67,548.2) -- (295.17,529.7) ;
\draw    (261.67,533.2) -- (281.67,533.2) ;
\draw    (266.67,528.2) -- (286.67,528.2) ;
\draw    (281.67,533.2) -- (281.67,553.2) ;
\draw    (286.67,528.2) -- (286.67,548.2) ;
\draw  [color={rgb, 255:red, 208; green, 2; blue, 27 }  ,draw opacity=1 ] (201.5,668.7) -- (210.5,668.7) -- (210.5,677.7) -- (201.5,677.7) -- cycle ;
\draw [color={rgb, 255:red, 208; green, 2; blue, 27 }  ,draw opacity=1 ]   (202,673.2) -- (211,673.2) ;
\draw [color={rgb, 255:red, 208; green, 2; blue, 27 }  ,draw opacity=1 ]   (206.5,668.2) -- (206,678.2) ;
\draw  [fill={rgb, 255:red, 255; green, 255; blue, 255 }  ,fill opacity=1 ] (189.23,616.86) -- (240.17,616.86) -- (240.17,643.26) .. controls (208.33,643.26) and (214.7,652.78) .. (189.23,646.62) -- cycle ; \draw  [fill={rgb, 255:red, 255; green, 255; blue, 255 }  ,fill opacity=1 ] (182.87,620.86) -- (233.8,620.86) -- (233.8,647.26) .. controls (201.97,647.26) and (208.33,656.78) .. (182.87,650.62) -- cycle ; \draw  [fill={rgb, 255:red, 255; green, 255; blue, 255 }  ,fill opacity=1 ] (176.5,624.86) -- (227.43,624.86) -- (227.43,651.26) .. controls (195.6,651.26) and (201.97,660.78) .. (176.5,654.62) -- cycle ;
\draw  [color={rgb, 255:red, 208; green, 2; blue, 27 }  ,draw opacity=1 ] (201.5,630.7) -- (210.5,630.7) -- (210.5,639.7) -- (201.5,639.7) -- cycle ;
\draw [color={rgb, 255:red, 208; green, 2; blue, 27 }  ,draw opacity=1 ]   (202,635.2) -- (211,635.2) ;
\draw [color={rgb, 255:red, 208; green, 2; blue, 27 }  ,draw opacity=1 ]   (206.5,630.2) -- (206,640.2) ;
\draw  [fill={rgb, 255:red, 255; green, 255; blue, 255 }  ,fill opacity=1 ] (189.23,529.7) -- (240.17,529.7) -- (240.17,552.24) .. controls (208.33,552.24) and (214.7,560.37) .. (189.23,555.11) -- cycle ; \draw  [fill={rgb, 255:red, 255; green, 255; blue, 255 }  ,fill opacity=1 ] (182.87,533.12) -- (233.8,533.12) -- (233.8,555.66) .. controls (201.97,555.66) and (208.33,563.79) .. (182.87,558.53) -- cycle ; \draw  [fill={rgb, 255:red, 255; green, 255; blue, 255 }  ,fill opacity=1 ] (176.5,536.53) -- (227.43,536.53) -- (227.43,559.08) .. controls (195.6,559.08) and (201.97,567.2) .. (176.5,561.94) -- cycle ;
\draw  [color={rgb, 255:red, 208; green, 2; blue, 27 }  ,draw opacity=1 ] (201.5,539.7) -- (210.5,539.7) -- (210.5,548.7) -- (201.5,548.7) -- cycle ;
\draw [color={rgb, 255:red, 208; green, 2; blue, 27 }  ,draw opacity=1 ]   (202,544.2) -- (211,544.2) ;
\draw [color={rgb, 255:red, 208; green, 2; blue, 27 }  ,draw opacity=1 ]   (206.5,539.2) -- (206,549.2) ;
\draw    (335.14,502) -- (95.14,504) ;
\draw    (95.14,504) -- (94.2,472) ;
\draw [shift={(94.14,470)}, rotate = 88.32] [color={rgb, 255:red, 0; green, 0; blue, 0 }  ][line width=0.75]    (10.93,-3.29) .. controls (6.95,-1.4) and (3.31,-0.3) .. (0,0) .. controls (3.31,0.3) and (6.95,1.4) .. (10.93,3.29)   ;

\draw (62,671) node [anchor=north west][inner sep=0.75pt]   [align=left] {Layer 1};
\draw (62,634) node [anchor=north west][inner sep=0.75pt]   [align=left] {Layer 2};
\draw (70,579.7) node [anchor=north west][inner sep=0.75pt]   [align=left] {......};
\draw (52, 707) node [anchor=north west][inner sep=0.75pt]   [align=left] {Embedding};
\draw (40,765) node [anchor=north west][inner sep=0.75pt]   [align=left] {$W_0$};
\draw (112.5,765) node [anchor=north west][inner sep=0.75pt]   [align=left] {$G$};
\draw (174.5,765) node [anchor=north west][inner sep=0.75pt]   [align=left] {$D$};
\draw (240.5,765) node [anchor=north west][inner sep=0.75pt]   [align=left] {$W_n$};
\draw (79,755.2) node [anchor=north west][inner sep=0.75pt]   [align=left] {...};
\draw (143,755.2) node [anchor=north west][inner sep=0.75pt]   [align=left] {...};
\draw (209,755.2) node [anchor=north west][inner sep=0.75pt]   [align=left] {...};
\draw (194.5,691.44) node [anchor=north west][inner sep=0.75pt]  [color={rgb, 255:red, 154; green, 47; blue, 47 }  ,opacity=1 ] [align=left] {\begin{minipage}[lt]{38.53pt}\setlength\topsep{0pt}
\begin{center}
{\tiny \textbf{Attention}}\\{\tiny \textbf{Head}}\\{\tiny \textbf{Weights}}
\end{center}

\end{minipage}};
\draw (190,579.7) node [anchor=north west][inner sep=0.75pt]   [align=left] {......};
\draw (258,579.7) node [anchor=north west][inner sep=0.75pt]   [align=left] {......};
\draw (111,447) node [anchor=north west][inner sep=0.75pt]   [align=left] {{\tiny Pooling}};
\draw (237,432) node [anchor=north west][inner sep=0.75pt]   [align=left] {Classifier};
\draw (190,668) node [anchor=north west][inner sep=0.75pt]  [color={rgb, 255:red, 208; green, 2; blue, 27 }  ,opacity=1 ] [align=left] {{\footnotesize $g$}};
\draw (201,678) node [anchor=north west][inner sep=0.75pt]  [color={rgb, 255:red, 208; green, 2; blue, 27 }  ,opacity=1 ] [align=left] {{\footnotesize $d$}};
\draw (190,630) node [anchor=north west][inner sep=0.75pt]  [color={rgb, 255:red, 208; green, 2; blue, 27 }  ,opacity=1 ] [align=left] {{\footnotesize $g$}};
\draw (201,640) node [anchor=north west][inner sep=0.75pt]  [color={rgb, 255:red, 208; green, 2; blue, 27 }  ,opacity=1 ] [align=left] {{\footnotesize $d$}};
\draw (62,538) node [anchor=north west][inner sep=0.75pt]   [align=left] {\begin{minipage}[lt]{36.74pt}\setlength\topsep{0pt}
\begin{center}
Layer L
\end{center}

\end{minipage}};
\draw (190,540) node [anchor=north west][inner sep=0.75pt]  [color={rgb, 255:red, 208; green, 2; blue, 27 }  ,opacity=1 ] [align=left] {{\footnotesize $g$}};
\draw (201,550) node [anchor=north west][inner sep=0.75pt]  [color={rgb, 255:red, 208; green, 2; blue, 27 }  ,opacity=1 ] [align=left] {{\footnotesize $d$}};

\end{tikzpicture}

\caption{Weights of attention heads of transformer LM as input to probing classifier. }
\label{fig:model}
\end{center}
\end{figure}

\begin{table*}[t]
\centering
  \scalebox{1}{
    \begin{tabular}{l|l|cccc|cccc}
      &             & \multicolumn{4}{c|}{$Dist>3$}                       & \multicolumn{4}{c}{$Dist>2$} \\
      & Model       &  $Acc$        & $P$         & $R$         &  $F1$     &  $Acc$        & $P$         & $R$         &  $F1$  \\ \hline
      \multirow{4}{*}{\STAB{\rotatebox[origin=c]{90}{Finnish}}}
      & LogReg      & 79.20 & {\bf 85.04} & 75.35 & 79.90 & 77.42 & 85.16 & 72.63 & 78.40  \\ 
      & MLP-1       & 78.85 & 82.56 & 77.91 & 80.17 & 78.53 & 84.16 & 76.31 & 80.04  \\ 
      & MLP-2       & 77.13 & 81.32 & 75.70 & 78.41 & 76.54 & 81.85 & 75.04 & 78.30  \\ 
      & RF          & {\bf 80.61} & 82.82 & {\bf 81.57} & {\bf 82.19} & {\bf 79.71} & {\bf 84.21} & {\bf 78.82} & {\bf  81.43} \\
      \hline
      \multirow{4}{*}{\STAB{\rotatebox[origin=c]{90}{Russian}}}
      & LogReg      & 80.63 & 83.10 & 78.46 & 80.72 & 77.94 & 84.72 & 73.23 & 78.56  \\ 
      & MLP-1       & {\bf 84.87} & {\bf 86.20} & 84.19 & {\bf 85.18} & {\bf 82.05} & {\bf 87.01} & 79.32 & {\bf 82.99}  \\ 
      & MLP-2       & 83.86 & 84.88 & 83.66 & 84.27 & 81.65 & 86.35 & 79.29 & 82.67  \\ 
      & RF          & 83.77 & 81.75 & {\bf 88.32} & 84.90 & 81.29 & 84.45 & {\bf 81.01} & 82.70\\
      \hline
    \end{tabular}}
  \caption{Overall performance of the classifiers}
  \label{tab:ALL_performance}
\end{table*}

\section{Government Probing Classifier}
\label{sec:experiment}


We build the probing classifier based on the BERT model, specifically on the weights of its attention heads, from each transformer layer.  Figure~\ref{fig:model} illustrates the input to the probing classifier.  From each attention head, we extract the weights for the governor verb and its argument and concatenate these weights into a vector.  The vector will contain $L \cdot A$ elements, where $L$ is the number of transformer layers and $A$ is the number of attention heads.\footnote{In the default BERT-base settings, $L=A=12$.}  We use the vector as a one-dimensional representation for the governor-argument pair, which encodes the syntactic relations between the governor and the argument.  This vector is the final input to the probing classifier. 

In Figure~\ref{fig:model}, the governor word is $G$ and its dependent is $D$.  Any word in a sentence---including the governor and its dependents---may be segmented into multiple sub-word tokens.\footnote{In the figure, the governor and governee are each shown with 2 tokens, but it could be more or fewer.}  
We use byte-pair encoding (BPE) to split words into sub-word tokens before input to the LM.
We denote by $g$ a member of the list of tokens of $G$, and by $d$ a segment of $D$.  From each head, we collect the weights for {\em all pairs} of tokens $(g, d)$, where $g \in G$ and $d \in D$.  

First, this gives us a series of matrices ($2 \times 2$ in the illustration in the figure), one for each head in each layer.  
These matrices are concatenated into a tensor of dimension $ ( L \cdot A )  \times 2 \times 2 $. 
Next, {\em for each head}, we max-pool all pairs of governor and dependent tokens (in the figure, $2 \times 2$ pairs), to obtain the {\em pooled} attention weight, which reflects the relation between the governor and its dependent, one for each head.  These pooled attention weights form a one-dimensional vector of size $( L \cdot A )$.  This vector is the input to the government classifier.  

We experiment by training four types of probing classifiers: logistic regression, multi-layer perceptron (MLP) with one fully-connected layer, MLP with two fully-connected layers, and Random Forests. 
More details on the hyper-parameters of each classifier can be found in Appendix~\ref{sec:hyperparameter}. 
We implement all four classifiers\comment{, including the two MLP classifiers,} using the \href{https://scikit-learn.org/}{Scikit-Learn Toolkit} and mostly follow their default hyper-parameter settings.  

For logistic regression, we set the maximum number of training iterations to 10000.  For the MLP 1-layer classifier, we use 144 neurons.  We use the same number of neurons for the first layer in the MLP 2-layer classifier, and 72 neurons for its second layer.  For the Random Forest classifier, we use 300 trees.

\section{Results and Discussion}
\label{sec:result}
We train four sets of classifier for each language and for each distance group, corresponding to four classifier types in Section~\ref{sec:experiment}. 
For each set of classifiers, we train 12 classifiers using the attention head weights only from the first $N$ transformer layers, with $N=1,2,...12$.  
The first classifier of each set is trained with attention head weights from only the first transformer layer, whereas the last one is trained with all attention head weights.  Therefore, we expect the last model, which is trained with all weights, should be the best performing one in each set of classifiers.  Since training involves some randomness, We repeat the process 5 times and average the results for more objective evaluation.

\subsection{Detecting Government Relations}

We first explore the overall performance of the probing classifiers. Table~\ref{tab:ALL_performance} shows the performance of the probing classifiers trained with all attention head weights, average over five repetitions.
All classifiers reach generally high scores, which are---within each language---quite close to each other.  The random forest classifier gives the highest score in Finnish, while the MLP-1 classifier performs the best in Russian for most metrics (except recall when $Dist>2$). 
All classifiers also show a very close performance within each distance group.  Except for recall in Russian, the difference among all metrics is not statistically significant, with p-values all above 0.05.  The p-value for recall in Russian is 0.04. 
Since the test and training data contain a balanced number of positive and negative instances, we can conclude that the probing classifiers can distinguish well between government vs.~non-government relations, with accuracy and $F_1$ over 80\% for both Finnish and Russian.

\begin{table*}[t]
\centering
\scalebox{0.99}{
\begin{tabular}{l|l|l|cccc|cccc}
& &              & \multicolumn{4}{c|}{$Dist>3$}                   & \multicolumn{4}{c}{$Dist>2$} \\
& & Model        & $Acc$ & $P$ & $R$ & $F1$                          & $Acc$ & $P$ & $R$ & $F1$ \\ \hline
\multirow{8}{*}{\STAB{\rotatebox[origin=c]{90}{Finnish}}}
& \multirow{4}{*}{\STAB{\rotatebox[origin=c]{90}{Near}}}        
& LogReg   & 83.20 & 84.83 & 85.42 & 85.12 & 83.15 & 83.86 & 88.63 & 86.18 \\
& & MLP-1  & 83.98 & {\bf 86.52} & 84.72 & 85.61 & 85.39 & 85.02 & 91.47 & 88.13 \\
& & MLP-2  & 79.69 & 83.82 & 79.17 & 81.43 & 83.71 & {\bf 85.25} & 87.68 & 86.45 \\
& & RF     & {\bf 84.38} & 82.10 & {\bf 92.36} & {\bf 86.93} & {\bf 85.67} & 84.48 & {\bf 92.89} & {\bf 88.49} \\ 
\cline{2-11}

& \multirow{4}{*}{\STAB{\rotatebox[origin=c]{90}{Far}}}        
& LogReg   & 80.08 & 91.21 & 67.48 & 77.57 & 78.18 & 92.31 & 68.39 & 78.57 \\
& & MLP-1  & 83.40 & 86.73 & {\bf 79.67} & 83.05 & 80.30 & 87.21 & {\bf 77.72} & 82.19 \\ 
& & MLP-2  & 80.08 & 83.78 & 75.61 & 79.49 & 76.36 & 84.85 & 72.54 & 78.21 \\
& & RF     & {\bf 85.48} & {\bf 91.51} & 78.86 & {\bf 84.72} & {\bf 81.52} & {\bf 93.42} & 73.58 & {\bf 82.32} \\ \hline

\multirow{8}{*}{\STAB{\rotatebox[origin=c]{90}{Russian}}}
& \multirow{4}{*}{\STAB{\rotatebox[origin=c]{90}{Near}}}        
& LogReg   & 81.62 & 80.48 & 85.80 & 83.06 & 70.59 & 81.18 & 65.34 & 72.40 \\
& & MLP-1  & {\bf 85.54} & {\bf 85.07} & 87.88 & {\bf 86.45} & {\bf 77.06} & {\bf 89.99} & 68.81 & {\bf 77.99} \\
& & MLP-2  & 84.32 & 83.99 & 86.66 & 85.30 & 76.67 & 87.99 & {\bf 70.06} & 78.01 \\
& & RF     & 82.65 & 78.82 & {\bf 91.55} & 84.71 & 72.98 & 83.71 & 67.37 & 74.65 \\ 
\cline{2-11}

& \multirow{4}{*}{\STAB{\rotatebox[origin=c]{90}{Far}}}        
& LogReg   & 86.74 & {\bf 93.32} & 79.48 & 85.85 & 82.75 & 92.72 & 75.00 & 82.93 \\
& & MLP-1  & {\bf 89.68} & 92.48 & 86.65 & {\bf 89.47} & {\bf 85.40} & {\bf 92.81} & 80.06 & 85.96 \\ 
& & MLP-2  & 89.06 & 90.96 & {\bf 87.02} & 88.95 & 85.11 & 91.47 & 80.88 & 85.85 \\
& & RF     & 87.99 & 90.65 & 85.04 & 87.76 & 88.99 & 92.06 & {\bf 87.88} & {\bf 89.92} \\ \hline

\end{tabular}}
\caption{Performance on detecting \textit{Near} vs.~\textit{Far} government between governor and complement. }
\label{tab:near-far}
\end{table*}

{\bf Near vs.~Far Arguments:}
To confirm the generality\comment{test the selectivity} of the probing classifiers, we explore the {\em best} performing models trained with all attention head weights from our experiments, and evaluate their performance on near vs.~far governees separately. 
These tests check that the probes learn to predict government, rather than focusing only on the adjacency of governors to their arguments. 
The results are in Table~\ref{tab:near-far}.
For Finnish, the majority of the models have higher F1 scores on detecting the government of the near than the far governees.
However, the difference between far and near in $Dist>3$ is not statistically significant.
(The p-values for all metrics are above 0.05, except for recall.)  

\comment{!!! JH START}
For Russian, the models score higher on detecting far governees than near ones, while for the Finnish models the scores are reversed.
There is a significant difference, with all p-values below 0.05 for both $Dist > 2$ and $Dist > 3$. 
This confirms that the probes are not selective for the adjacency of the governors and their arguments.

\comment{!!! JH END}

\begin{figure*}
\begin{minipage}{0.49\linewidth}
\centering
  \includegraphics[scale=0.55, trim=12mm 7mm 15mm 13mm, clip ]{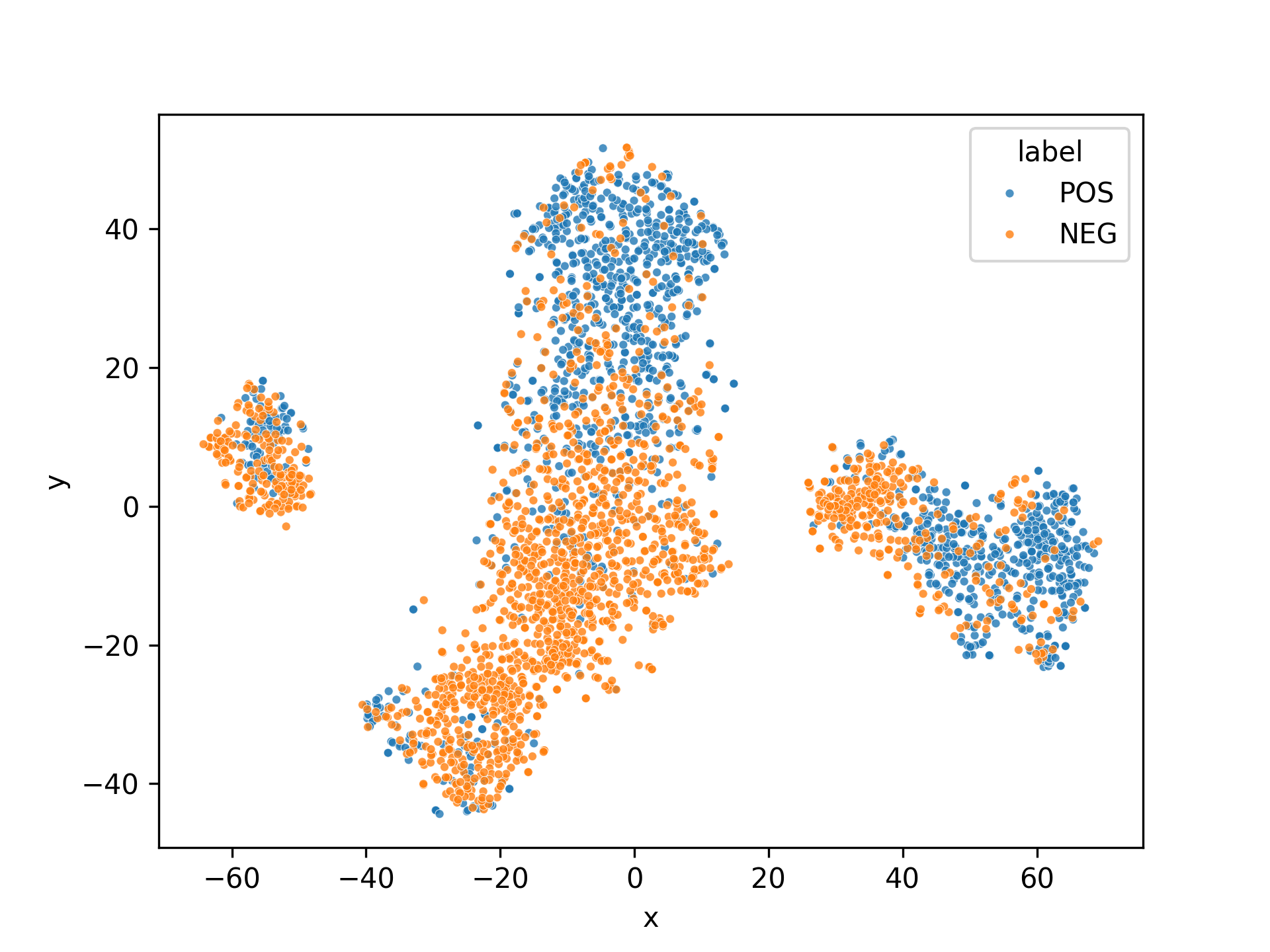}
\end{minipage}\hfill
\begin{minipage}{0.49\linewidth}
\centering
  \includegraphics[scale=0.55, trim=12mm 7mm 15mm 13mm, clip ]{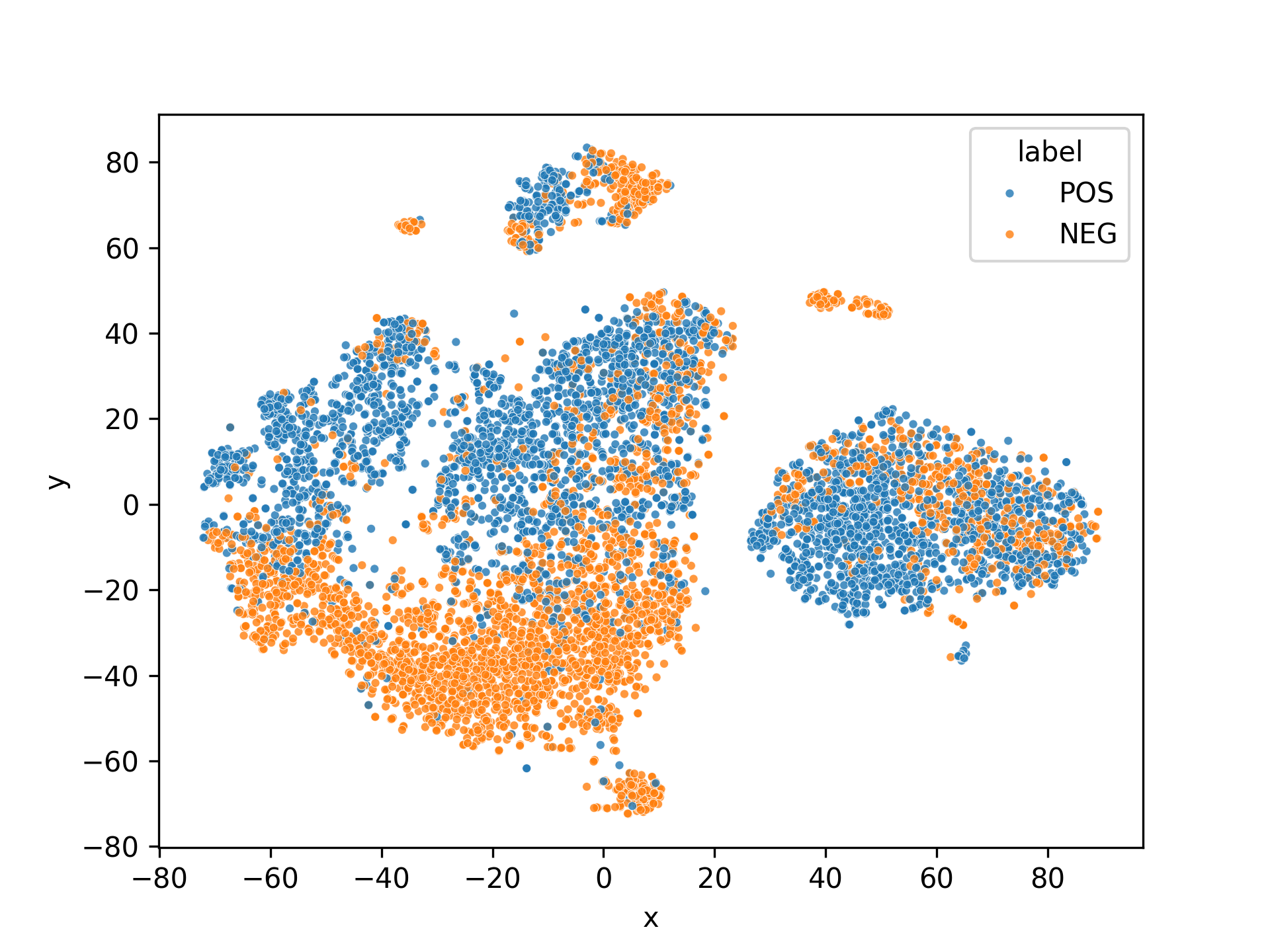}
\end{minipage}
\caption{t-SNE visualization of positive vs. negative instances. (left: Finnish, right: Russian)}
\label{fig:tsne}
\end{figure*}

{\bf Positive vs. Negative instances:}  To visualize the difference between the attention weights of the positive vs.~negative instances, we use the t-SNE dimensionality reduction technique~\cite{JMLR:v9:vandermaaten08a}, implemented in Scikit learn, with default parameters.  Figure~\ref{fig:tsne} shows how t-SNE projects the high-dimensional input vectors of the probing classifiers onto the 2-D plane.

We plot all instances from the test data, for both languages.\comment{!!!+++ Add details here and interpretation}
We can observe some strikingly well-defined clusters of the positive vs.~the negative instances, particularly for Russian.  This separability may help explain the very high performance of all probes on government prediction in Table~\ref{tab:ALL_performance}.


\begin{figure*}
\begin{minipage}{0.49\linewidth}
\centering
  \includegraphics[scale=0.5, trim=5mm 5mm 0mm 3mm, clip ]{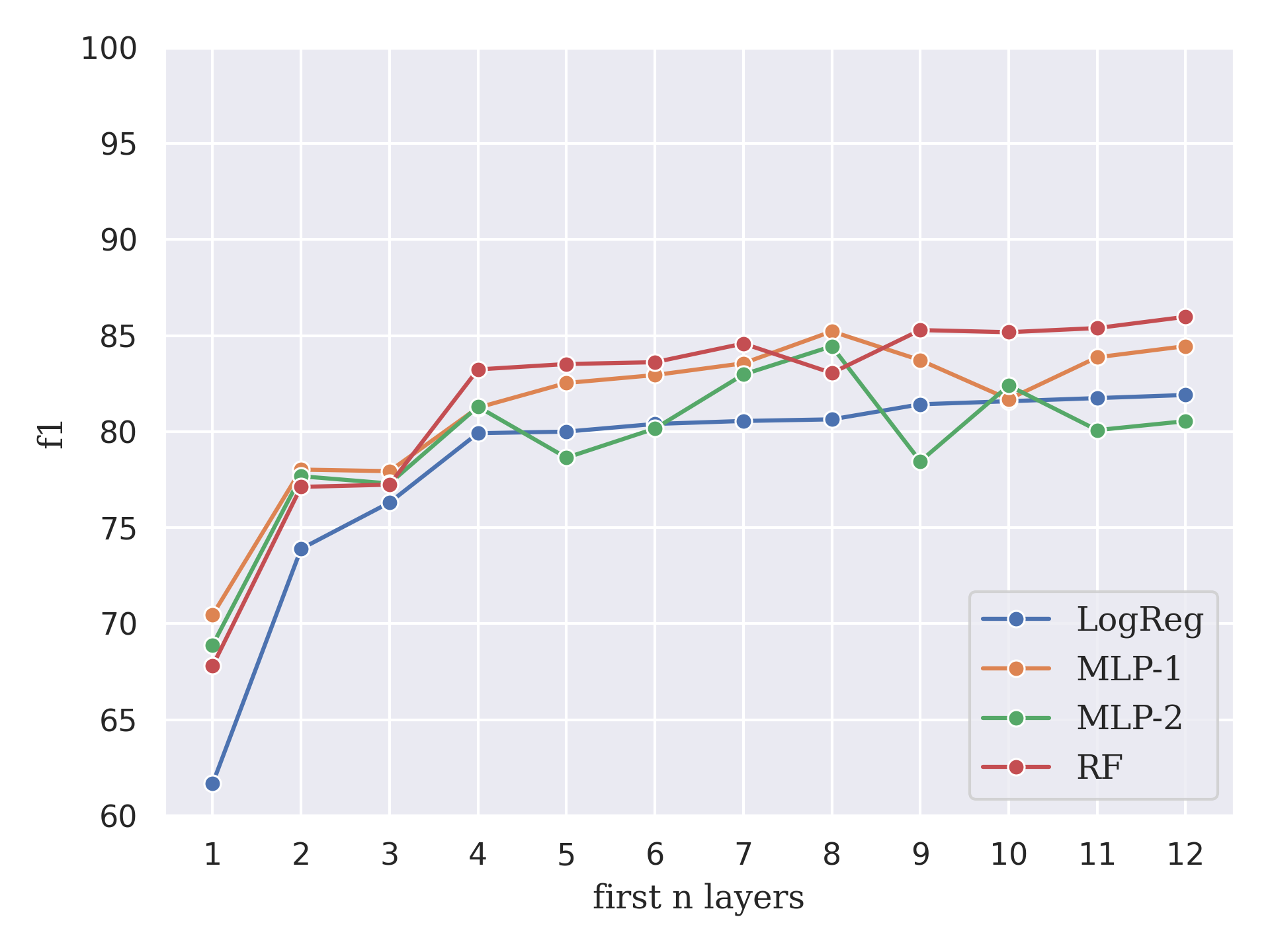}
\end{minipage}\hfill
\begin{minipage}{0.49\linewidth}
\centering
  \includegraphics[scale=0.5, trim=5mm 5mm 0mm 3mm, clip ]{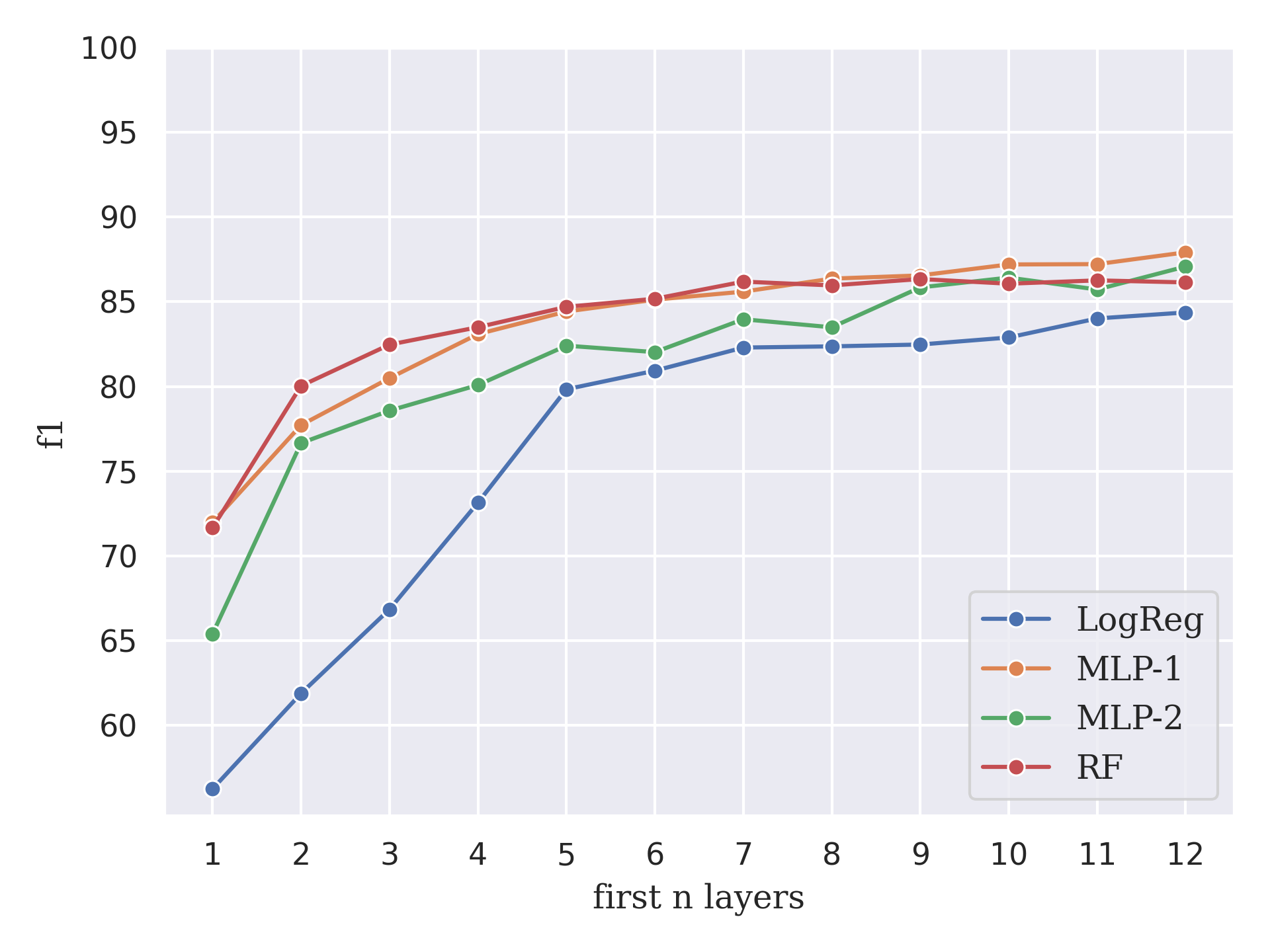}
\end{minipage}
\caption{Probing government prediction with attention weights from the first N layers of BERT (X-axis) when $Dist>3$. Y-axis---$F_1$ measure. (left: Finnish, right: Russian)}
\label{fig:layer_perf_dist_3}
\end{figure*}

\begin{figure*}
\begin{minipage}{0.49\linewidth}
\centering
  \includegraphics[scale=0.5, trim=5mm 5mm 0mm 3mm, clip ]{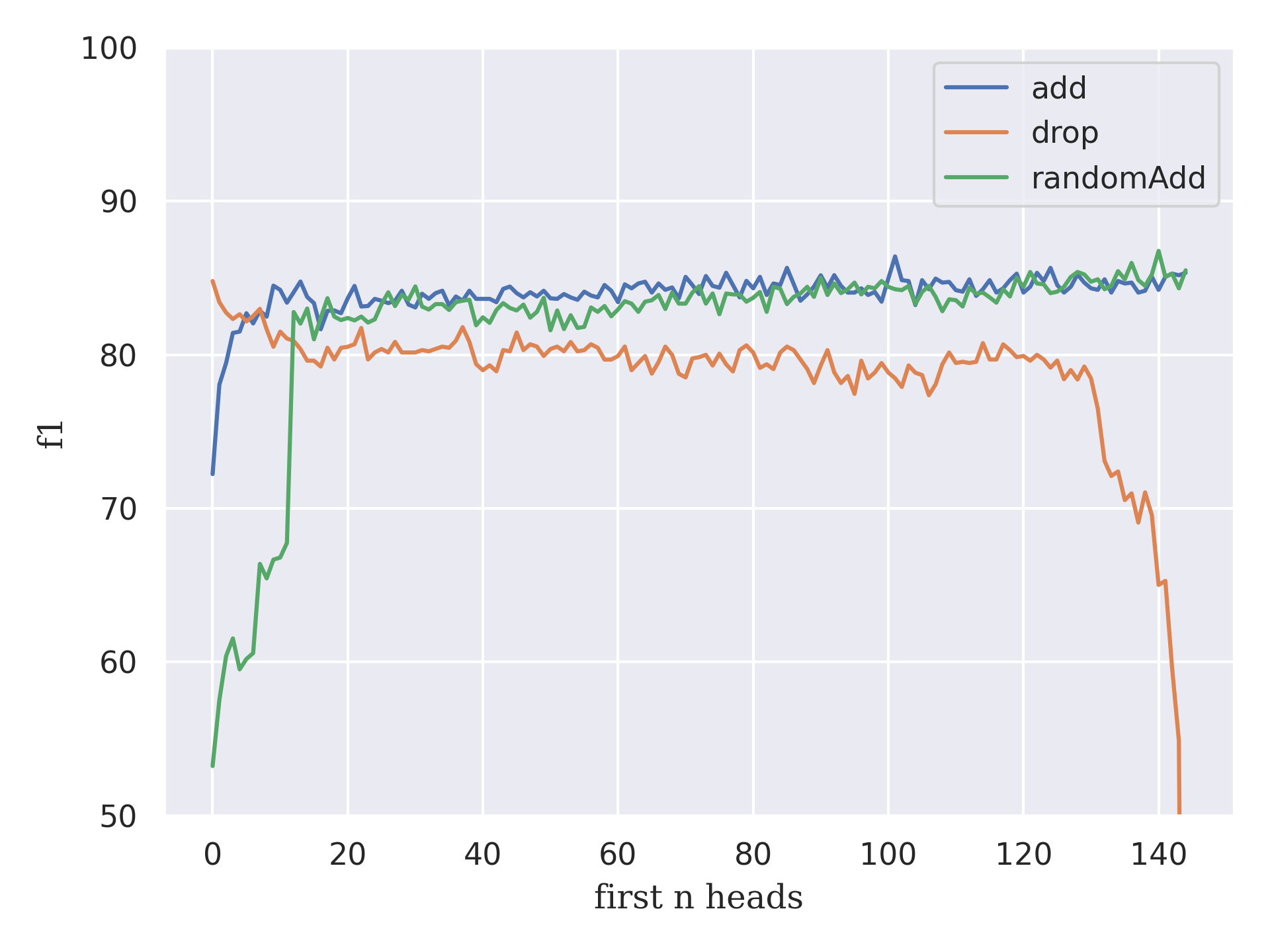}
\end{minipage}\hfill
\begin{minipage}{0.49\linewidth}
\centering
  \includegraphics[scale=0.5, trim=5mm 5mm 0mm 3mm, clip ]{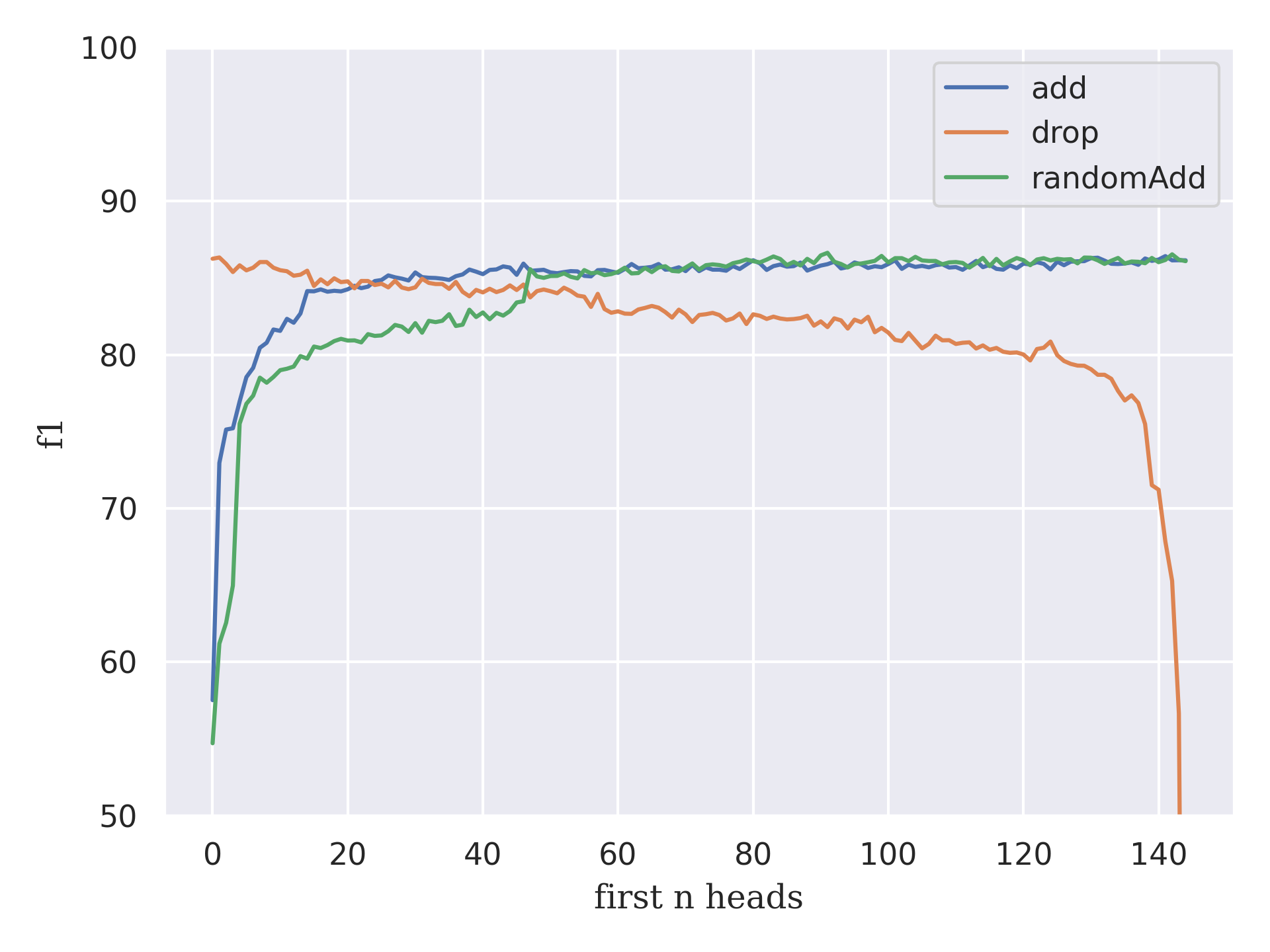}
\end{minipage}
\caption{$F_1$ score for Random forest classifier with selected attention heads when $Dist>3$ (left: Finnish, right: Russian).}
\label{fig:random_forest_n_heads_dist_3}
\end{figure*}

\subsection{Selection of Transformer Layers}

\comment{!!! JH START}
We next investigate how syntactic information about government is distributed across different transformer layers.  
We select the best-performing model trained with all attention weights, for $Dist > 3$, and visualize it together with all the other models in its classifier set.  This lets us evaluate the contribution of the first $N$ layers to the probe's ability to predict government.
\comment{!!! JH END}

Figure~\ref{fig:layer_perf_dist_3} shows the performance in terms of the $F_1$ measure, for all four classifier types.  Almost all curves increase monotonically as the number of selected layers increases, which is expected, as more information can be inferred with more transformer layers.
Crucially, all curves also indicate that the performance of classifiers increases rapidly for the early layers--layers 1 to 4 for Finnish, to layer 5--6 for Russian---and then the performance plateaus, with much smaller subsequent increases.  This suggests that the syntactic information important for government predictions is encoded in the {\em lowest} levels of BERT.  For Russian, this information is slightly more spread out across the first 5--6 layers. This mostly aligns with the observation in other probing contexts~\cite{hewitt-manning-2019-structural, lin-etal-2019-open, liu-etal-2019-linguistic, goldberg2019assessing}.

\comment{!!! JH START}
We check similar plots for models with $Dist > 2$ in Figure~\ref{fig:layer_perf_dist_2}, which can be found in Appendix~\ref{sec:plots_dist_2}, and which shows a very similar pattern as in Figure~\ref{fig:layer_perf_dist_3}.
\comment{!!! JH END}

\subsection{Ablation of Attention Heads}
\label{sec:ablation-studies}

We perform detailed ablation studies to see how attention heads at different layers contribute to government prediction.  We use the logistic regression classifiers, trained as explained above, and explore how the weights learned in their attention heads impact on each head's significance for government prediction.  
We fist rank all attention heads according to their logistic regression coefficients, and then ablate attention heads from two opposite perspectives: (A) training the classifier by {\em including} only the top-$N$ heads, and (B) training the classifier {\em excluding} the top-$N$ heads, with  $N$ ranging from $1$ to $144$ ($= 12 \times 12$).  For comparison, we also train classifiers using a {\em random} subset of $N$ heads, as a baseline.  
\comment{!!! JH START}

In this experiment, we assess the relative importance and contribution of the attention heads, by ``probing'' our probing classifiers.  We train the random forest classifier with $N=1,2,...144$, for both languages. We explore models with $Dist>3$ and plot their $F_1$ scores in Figure~\ref{fig:random_forest_n_heads_dist_3}. 

Models with $Dist>2$ show very similar behavior---we visualize their performance in Figure~\ref{fig:random_forest_n_heads_dist_2}, in Appendix~\ref{sec:plots_dist_2}.

\comment{!!! JH END}
We observe that for Finnish, the model achieves an $F_1$ score of 80\% either with the 3 top heads (represented by the blue curve), or by using all other heads together except the 3 top heads (orange curve).  This suggests that the top 3 heads contain most of the needed information and are reliable indicators of government.  Training without the top $N$ heads yields superior performance compared to the baseline, where the heads are chosen at random (represented by the green curve).

Similarly for Russian, 17--20 top heads are required to achieve the same performance ($F_1=~85$\%) as when excluding these top heads.  
This indicates that Russian BERT requires more heads to sufficiently represent government relations.  Training probing classifiers without the top $N$ heads eventually shows a slightly worse performance for Finnish, which indicates that the Finnish BERT has less distributed government information across its heads. 
It also suggests that the top $N$ heads are reliable, but {\em not the only} indicators of government relations between words.  Government can still be inferred well from all of the remaining attention heads as well.

\subsection{Error Analysis}

We manually examined\comment{all} instances that were misclassified by the probes. 
Some classification errors are caused by errors in the data (``{\em noise}'')---parsers for both languages occasionally assign incorrect dependency relations.  For example, in Finnish:
\begin{itemizerCompact}
\item[] ``Yritykset {\bf investoivat} eli laajentavat \underline{toimintaa}, rakentavat uutta ja hankkivat tekniikkaa.'' (\textit{Companies invest or expand operations, build new ones and acquire technology.})     
\end{itemizerCompact}
The parser incorrectly labels ``toimintaa'' (\textit{operations}) as the object of the governor ``investoida'' (\textit{invest}).  In fact, it is governed by another verb, ``laajentaa'' (\textit{expand}), which is conjoined with ``investoivat''.   All probes correctly predicted ``no government,''  which decreases recall in our evaluation.

Similar problems occur in Russian data:
\begin{itemizerCompact}
    \item[] 
    \textcyr{``Причем больше \underline{всего} на свете они \textbf{боялись}, что кто-нибудь о нем узнает.''} \\
    (\textit{Incidentally, more than anything, they feared that someone would find out about it.}) 
\end{itemizerCompact}
The parser labels ``\textcyr{всего}'' (\textit{everything}) as an object of the governor ``\textcyr{бояться}'' (\textit{fear}).  In reality, this verb's object is the following relative clause, and ``\textcyr{больше всего}'' (\textit{more than anything}) is an adverbial phrase modifier.

There are some inconsistencies between the representation of dependency relations---on one hand by, the third-party neural parsers, and and on the other hand, by other components that we use to detect positive vs.~negative government instances.  This leads to some amount of misclassified instances.  For example, in Finnish, subjects of passive verbs in the nominative case are parsed as objects.  While the probing classifiers {\em correctly} reject government relations in such instances, this results in ``false negatives'' in evaluation.  These examples represent ``noise'' in the data.  They will be fixed in future work; this gives us hope that after that the results may further slightly improve.

Some proportion of the instances are true misclassifications.  An analysis of these errors reveals that, occasionally, the probes fail to detect government relations with a long-distance governee that {\em precedes} its governor, or incorrectly identify adjacent adverbial modifiers as arguments.  In Finnish, true misclassifications are rare, mostly caused by missed arguments and misclassified adjacent words.

\subsection{Discovering New Government Patterns}

Research question RQ2 is: Are the probing classifiers capable of discovering new, previously unseen government patterns?  To evaluate the generalization power of the probing classifiers, and check whether they can be used to enhance the Government Bank, we assess the performance of the probes on governors and government patterns that were held out from the training data (``unseen''). 

{\bf Unseen government patterns:} For Finnish, we ran each probing classifier 8 times, while withholding two patterns from the labeled data, using these patterns only in the test data.  In each run, we held out different patterns, e.g., all arguments in ablative case, or all third-infinitive arguments (e.g., ``Hän \textbf{auttoi} \underline{leipomaan} leipää.'', ``\textit{S/he \textbf{helped} \underline{to bake} bread.}''). 
We test analogously for Russian. 

Table~\ref{tab:perf_govp} presents the accuracy of discovering unseen government. The overall good performance implies that we can use the information encoded in the LM to discover new patterns, and extend the Government Bank. Higher accuracy for the Random Forest classifier suggests that the probes are capable of finding more unseen patterns compared to other classifiers.  

\comment{We performed 5 runs (for each probing model) for Russian by removing patterns where an argument is a prepositional phrase, with different prepositions, e.g., any preposition governing genitive case or preposition ``\textcyr{о}'' (\textit{about}) governing locative case.}

\begin{table}[t]
\centering
\begin{tabular}{l|l|r|r}
& Model         & Finnish & Russian  \\ \hline
\multirow{4}{*}{\STAB{\rotatebox[origin=c]{90}{$Dist>3$}}}
& LogReg       & 79.83 & 80.63    \\ 
& MLP-1   & 80.17 & 84.87    \\
& MLP-2   & 79.83 & 83.86    \\
& RF & \textbf{83.79} & \textbf{83.77}     \\ \hline
\multirow{4}{*}{\STAB{\rotatebox[origin=c]{90}{$Dist>2$}}}
& LogReg       & 69.79 & 77.94    \\ 
& MLP-1   & 73.81 & 82.05    \\
& MLP-2   & 72.99 & 81.65    \\
& RF & \textbf{77.42} & \textbf{81.29}     \\ \hline
\end{tabular}
\caption{Accuracy of classifiers on discovering unseen government patterns for Finnish and Russian. Reported metrics are micro-averaged.}
\label{tab:perf_govp}
\end{table}

{\bf Unseen governing verbs: } We perform a similar assessment while withholding verbs from the training data.  For each language, we run the probing classifiers 5 times.  For each run, we withhold a random set of 66 verbs for Finnish (146 verbs for Russian).  This lets us check the ability of the probes to discover new, unseen governor verbs.

Table~\ref{tab:governor-unseen} shows the performance, averaged across all runs.  The results indicate that probing classifiers can detect new verbs governing their complements, not previously seen during training. 

These evaluations conclusively demonstrate the practical applicability of the government probing results---the information encoded in BERT's attention weights can be used for building new valuable resources---Government Banks.

\begin{table}[t]
  \scalebox{0.98}{
    \begin{tabular}{l|l|rr|rr}
      &              & \multicolumn{2}{c|}{$Dist>3$}                   & \multicolumn{2}{c}{$Dist>2$} \\
      & {\em Model}    &  $Acc$    & $F1$          & $Acc$          & $F1$ \\ \hline
      \multirow{4}{*}{\STAB{\rotatebox[origin=c]{90}{Finnish}}}     
      & LogReg         & 79.12  & 78.15  & 78.86  & 77.53       \\ 
      & MLP-1          & 78.67  & 78.52  & 79.54  & 78.90       \\ 
      & MLP-2          & 76.82  & 76.49  & 77.35  & 76.83       \\ 
      & RF             & {\bf 79.93}  & {\bf 80.12}  & {\bf 79.93}  & {\bf 79.50} \\
      \hline
      \multirow{4}{*}{\STAB{\rotatebox[origin=c]{90}{Russian}}} 
      & LogReg        & 80.63  & 80.72  & 77.46  & 77.69       \\ 
      & MLP-1         & {\bf 84.87}  & {\bf 85.18}  & {\bf 81.83}  & {\bf 82.49}       \\ 
      & MLP-2         & 83.86  & 84.27  & 81.41  & 82.14 \\
      & RF            & 83.77  & 84.90  & 80.86  & 82.01       \\
      \hline
    \end{tabular}}
  \caption{Performance of classifiers on unseen governors.  Reported metrics are micro-averaged.}
  \label{tab:governor-unseen}
\end{table}

\section{Conclusion}
\label{sec:conclusion}


The contribution of this paper is twofold:  \\
1.~We release a Government Bank (for Finnish and Russian) to the research community for work on the representation of grammatical government and constructions in neural language models. \\
2.~We present an exploration of the representation of government constructions in transformer LMs.  To the best of our knowledge, these are the first such resources to be made publicly available---not only in human-readable, but also in machine-usable form; and the first exploration of how well LMs can predict complex constructions, such as government.

We probe the syntactic information encoded in BERT's attention heads, to reveal what it knows about government relations.  
Our objective is to extract knowledge about government relations from inside the LM's attention mechanism.  

We evaluate the performance of the classifiers on Finnish and Russian data, from several perspectives.  To study RQ1, we assess the overall performance of the classifiers, and explore their probing selectivity with respect to the distance between the governor and its governees.  We show that the performance of the probing classifiers is very high.  Notably, the classifiers perform as well or better when the governee is far from its governor.  We probed for the distribution of information about government across BERT's attention heads across different layers.  We show that the probing classifiers are able to infer sufficient information from the first few layers of BERT.

We further explore the contribution of each attention head, using the coefficients learned by the logistic regression classifier.  The idea is that the learned coefficients serve as a good indication of the contribution of the corresponding heads.  Overall, the result shows that the most ``important'' attention heads (with the highest coefficients) are reliable indicators of government relations, but they are not the only ones.  The probing classifiers are capable of detecting government using the remaining heads, which have lower coefficients.  Comparing Finnish and Russian, we show that the information is slightly less spread out across the Finnish BERT heads, as compared to Russian BERT heads.  We also performed an error analysis to understand the limitations of our probing classifiers.

For RQ2, we evaluate the ability of the probe to identify {\em novel} government relations, never seen during the training phase.  We held out: (a) specific government patterns, and (b) governing verbs.  The experiments show that the probes are able to discover novel government relations and novel governors, unseen in training. 

In future work, we plan to use larger datasets, and identify large sets of government relations.  This will extend the Government Bank with new patterns.
We plan to work with additional languages, in particular, we are extending this approach to German and Italian government.
Crucially, we will extend this work government of other parts of speech (nouns, adjectives, etc.), and to more complex types of constructions than government.

\section*{Limitations}

The current work has a number of limitations to consider.
(A) For now, our probing of government relations is limited to two languages.  Extending to additional languages is challenging when probing this type of construction because it requires a Government Bank for each additional language, and collection of language-specific data.  We are currently conducting experiments with an Italian and a German Government Bank; if the paper is accepted, we expect to be able to include those results as well.

(B) So far, we have performed probing using only one type of transformer models---BERT-base.  In future work, we plan to extend experiments to other models as well.

(C) So far, we have performed only one type of probing---using correlation probing with probing classifiers.  This type of probing has received some criticism, of which we are aware.  In future work, we experiment with the model representations from hidden layers and other types of probing methodologies.

\section*{Ethics Statement}

We do not see any ethical issues with the current work.  We use publicly available resources for all conducted experiments, and release language resources, which were created in collaboration with linguists, who were aware of how the data will be used.

\section*{References} 

\bibliographystyle{lrec-coling2024-natbib}
\bibliography{references} 



\begin{table}[t]
\centering
\begin{tabular}{l|l|r}
{\em PoS}                     & {\em Feature / Case}        & {\em total} \\ \hline
\multirow{9}{*}{Noun}        & Inessive  & 2509  \\
                             & Genitive  & 2424  \\
                             & Illative  & 2337  \\
                             & Elative  & 2190  \\
                             & Adessive  & 1630  \\
                             & Essive  & 1529  \\
                             & Allative  & 1096  \\
                             & Ablative  & 522  \\
                             & Translative  & 659  \\ \hline
\multirow{2}{*}{Verb}         & Ma-Infinitive    & 1112     \\
                              & A-Infinitive       & 1555   \\ \hline
\multirow{2}{*}{Postposition} & Partitive   & 2        \\
                              & Genitive    & 11        \\ \hline
\multirow{9}{*}{Adjective}   &Ablative & 93  \\
                             &Adessive & 65  \\
                             &Allative & 42  \\
                             &Elative & 77  \\
                             &Essive & 230  \\
                             &Genitive & 312  \\
                             &Illative & 35  \\
                             &Inessive & 54  \\
                             &Translative & 235  \\  \hline
total                         &             & 18719 
\end{tabular}
\caption{Detailed overview of Finnish instances, before sampling}
\label{tab:fin_instances}
\end{table}

\begin{table}[t]
\centering

  \begin{tabular}{l|l|l|r}
{\em PoS}                      & \multicolumn{2}{l|}{\em Feature / Case}        & {\em total} \\ \hline
Verb                           & \multicolumn{2}{l|}{Infinitive}         & 41090  \\ \hline
\multirow{3}{*}{\STAB{\rotatebox[origin=c]{90}{Noun}}} 
                               & \multicolumn{2}{l|}{Dative}            & 11406  \\
                               & \multicolumn{2}{l|}{Genitive}           & 28926   \\
                               & \multicolumn{2}{l|}{Instrumental}       & 8671  \\ \hline
\multirow{26}{*}{\STAB{\rotatebox[origin=c]{90}{Preposition}}}  & \textcyr{в}      & \multirow{5}{*}{Accusative}   & 6674   \\
                               & \textcyr{за}     &                               & 1520   \\
                               & \textcyr{на}     &                               & 8599   \\
                               & \textcyr{о}      &                               & 42     \\
                               & \textcyr{под}    &                               & 17    \\
                               & \textcyr{про}      &                               & 236   \\   \cline{2-4} 
                               & \textcyr{к}      & \multirow{2}{*}{Dative}       & 4278   \\
                               & \textcyr{по}     &                               & 933    \\ \cline{2-4} 
                               & \textcyr{без}    & \multirow{10}{*}{Genitive}    & 54     \\
                               & \textcyr{для}    &                               & 210    \\ 
                               & \textcyr{до}     &                               & 480    \\
                               & \textcyr{из}     &                               & 2111   \\
                               & \textcyr{из-за} &                               & 25     \\ 
                               & \textcyr{из-под} &                               & 16     \\ 
                               & \textcyr{от}     &                               & 2598   \\
                               & \textcyr{против} &                              & 56     \\
                               & \textcyr{ради}   &                               & 6     \\
                               & \textcyr{с}      &                               & 1150   \\
                               & \textcyr{у}      &                               & 800    \\ \cline{2-4} 
                               & \textcyr{за}     & \multirow{6}{*}{Instrumental} & 840    \\
                               & \textcyr{между}  &                               & 20     \\
                               & \textcyr{над}    &                               & 234    \\
                               & \textcyr{перед}  &                              & 38     \\
                               & \textcyr{под}    &                                & 7     \\
                               & \textcyr{с}      &                              & 3477   \\ \cline{2-4} 
                               & \textcyr{в}      & \multirow{3}{*}{Locative}     & 3925   \\
                               & \textcyr{на}     &                              & 2425  \\
                               & \textcyr{о}      &                               & 2816   \\ \hline
                               &  Total      &                              & 143836
\end{tabular}
\caption{Detailed overview of Russian instances, before sampling}
\label{tab:rus_instances}
\end{table}

\section*{Appendices} %
\appendix

\section{Overall instance statistics}
\label{sec:instance_stat}
\comment{!!! JH START}
Table~\ref{tab:fin_instances} and Table~\ref{tab:rus_instances} show the distribution of different types of governees---syntactic dependents of the head verb, which are governed by the verb---before balancing and sub-sampling.  
We should mention that we exclude certain common governees from our experiments, despite the fact that they are present in our government bank.  The following direct objects governees are excluded: for Finnish---nouns in objective case and in partitive case, and for Russian noun in the accusative case.  This is done because we are unable to find negative instances with these cases for data balancing, unlike for all other governee types.
A large proportion of ``negative'' instances for these cases have been manually verified as positive instances.  This suggests that our government bank is not yet complete, which is as expected.  Therefore, we exclude these types of governees from our experiments.
\comment{!!! JH END}

\section{Hyper-parameters of classifiers}
\label{sec:hyperparameter}

We implement the classifiers, including both MLP 1-layer and 2-layer classifiers, based on the Scikit Learn library,\footnote{https://scikit-learn.org/} and mostly follow their default hyper-parameter settings. 

For the logistic regression classifier, we set the maximum number of training iterations to $10000$.  

For the MLP 1-layer classifier, we set $144$ neurons.  We use the same number of neurons for the first fully-connected layer in the MLP 2-layer classifier, while we set its second fully-connected layer to $72$ neurons.  

For the Random Forest classifier, we use $300$ trees.

\section{Selection of Transformer Layers and Attention Heads for $Dist>2$}
\label{sec:plots_dist_2}

\begin{figure*}[h]
\begin{minipage}{0.49\linewidth}
\centering
  \includegraphics[scale=0.5, trim=5mm 5mm 0mm 3mm, clip ]{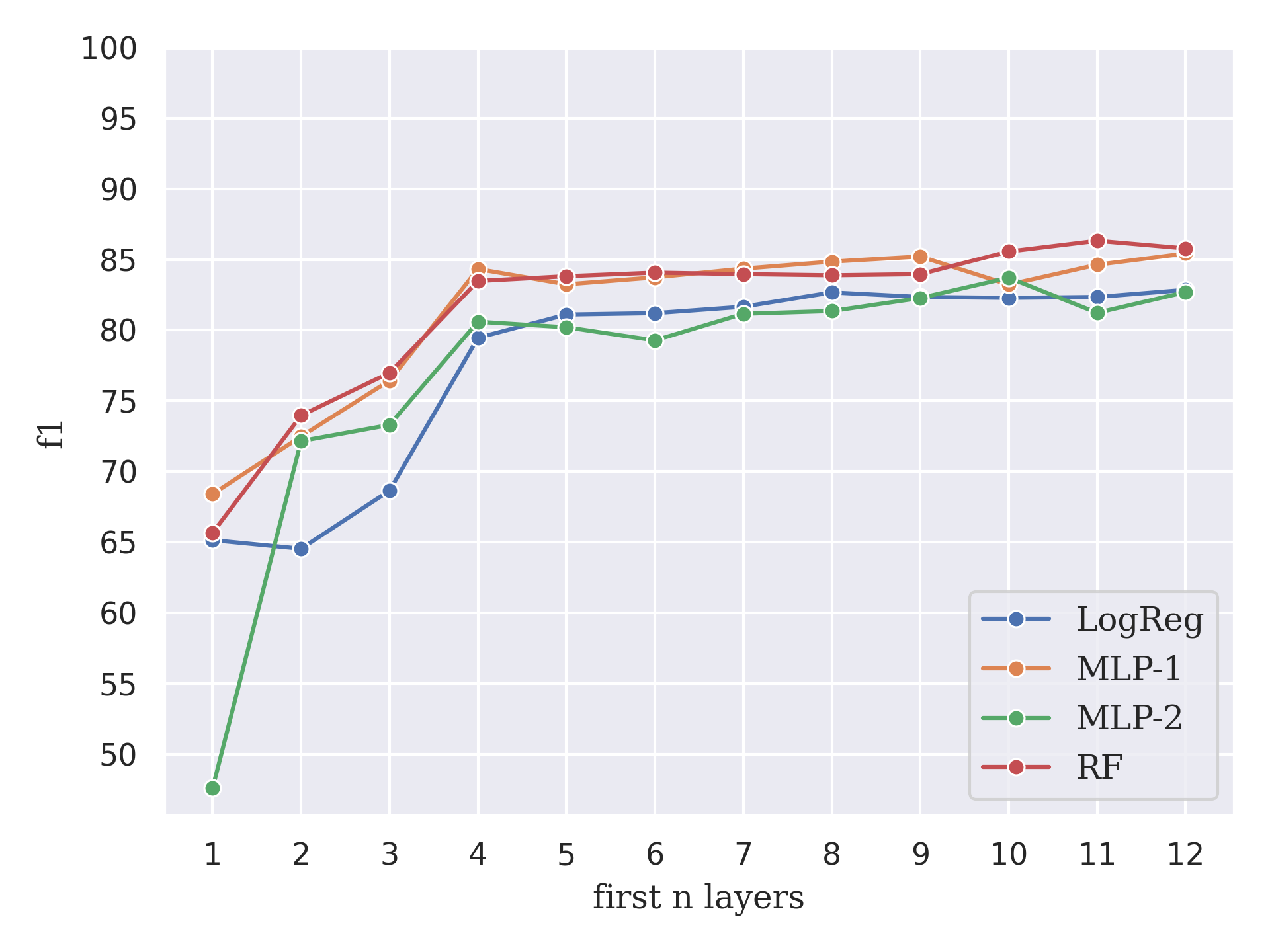}
\end{minipage}\hfill
\begin{minipage}{0.49\linewidth}
\centering
  \includegraphics[scale=0.5, trim=5mm 5mm 0mm 3mm, clip ]{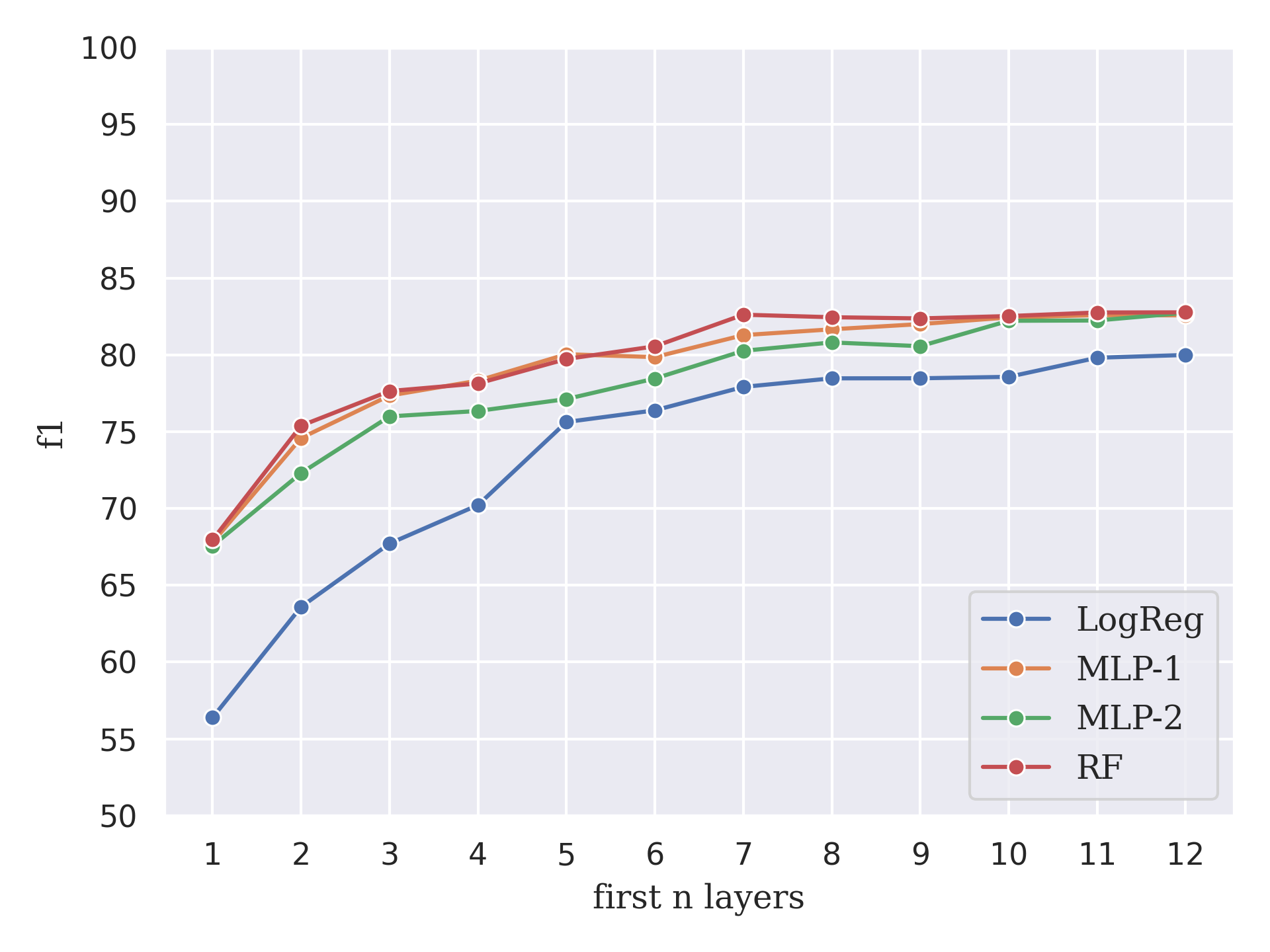}
\end{minipage}
\caption{Probing government prediction with attention weights from the first N layers of BERT (X-axis) when $Dist>2$. Y-axis---$F_1$ measure. (left: Finnish, right: Russian)}
\label{fig:layer_perf_dist_2}
\end{figure*}

\begin{figure*}
\begin{minipage}{0.49\linewidth}
\centering
  \includegraphics[scale=0.5, trim=5mm 5mm 0mm 3mm, clip ]{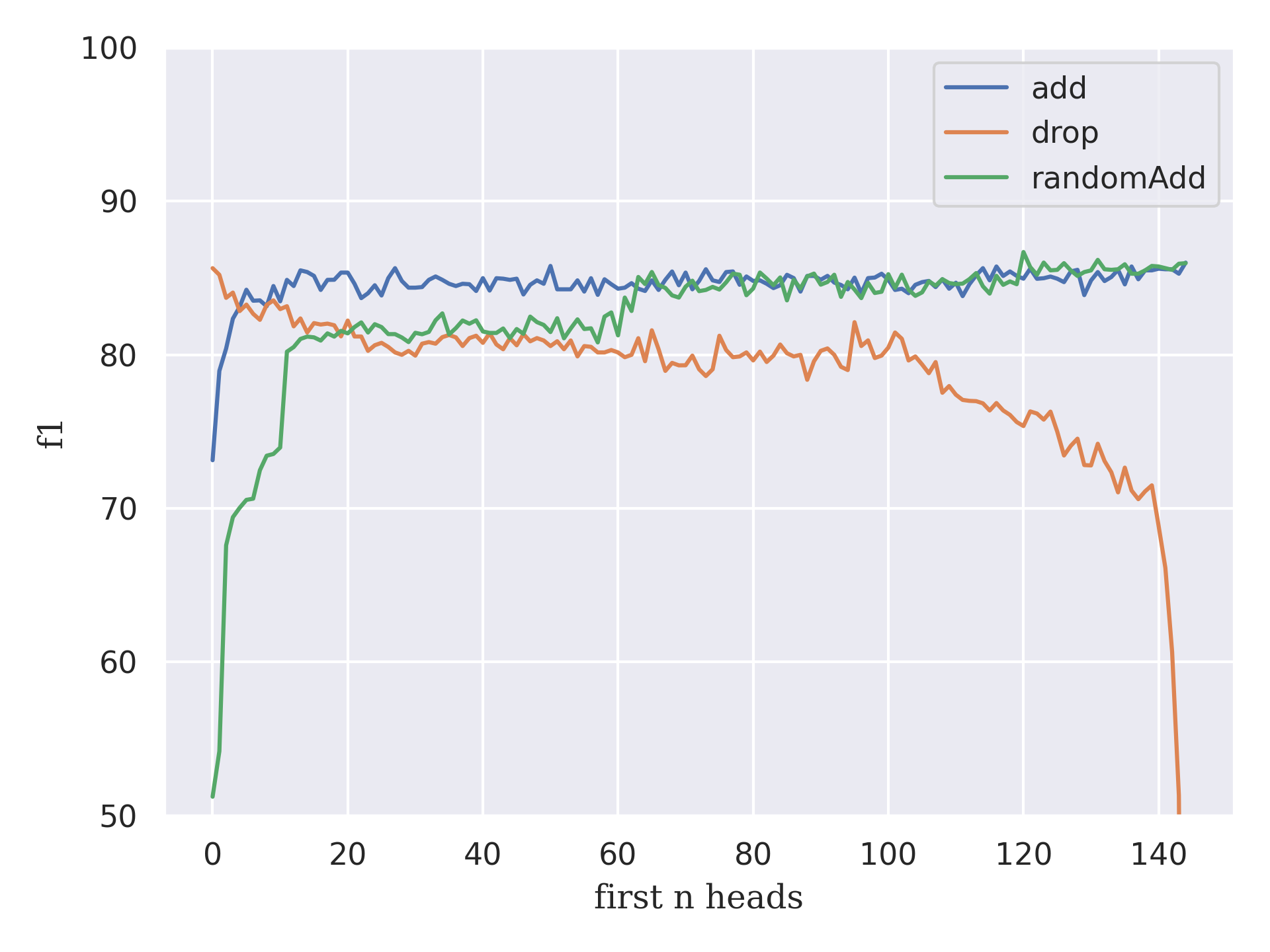}
\end{minipage}\hfill
\begin{minipage}{0.49\linewidth}
\centering
  \includegraphics[scale=0.5, trim=5mm 5mm 0mm 3mm, clip ]{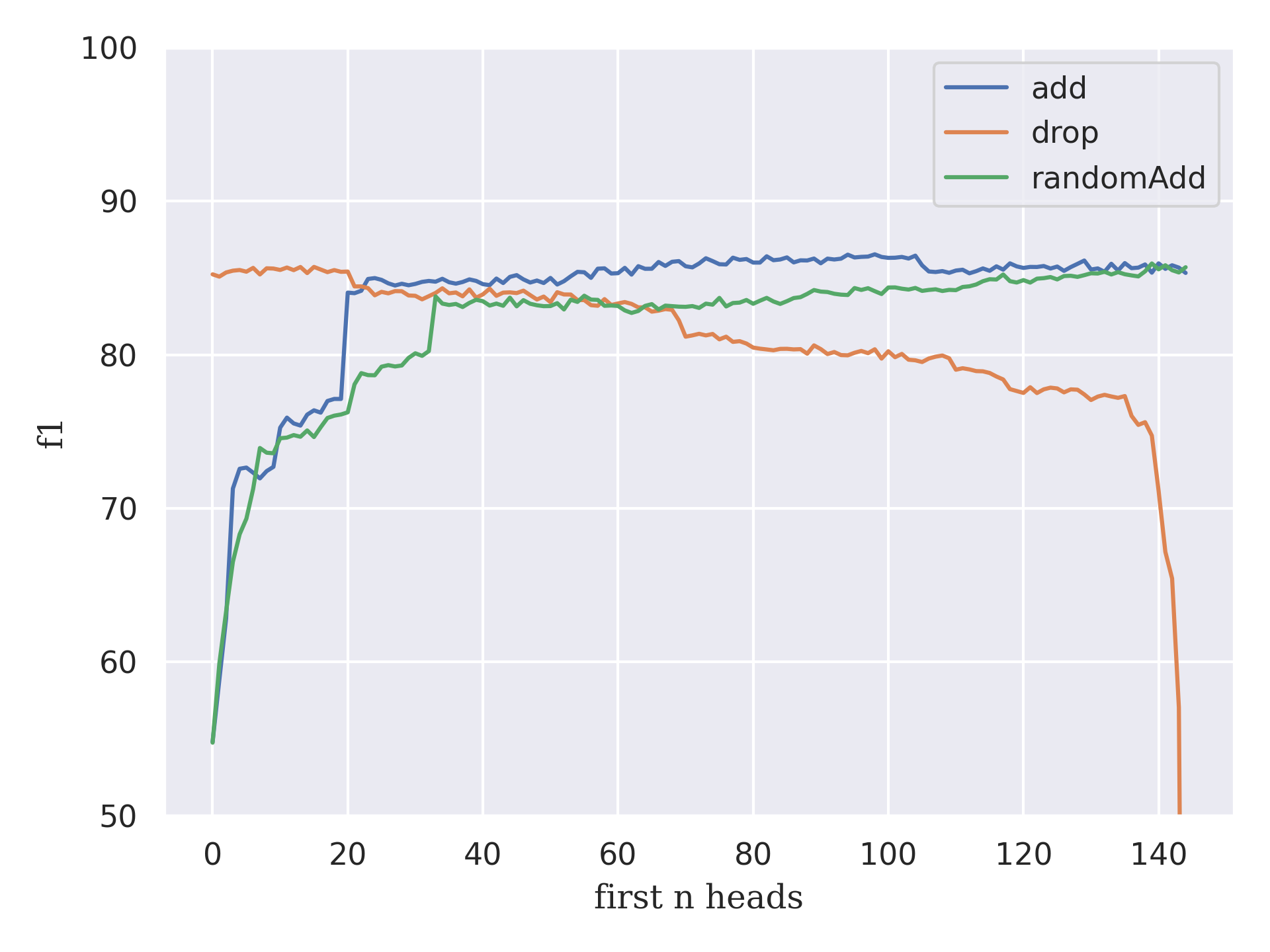}
\end{minipage}
\caption{$F_1$ score for Random forest classifier with selected attention heads when $Dist>2$ (left: Finnish, right: Russian.}
\label{fig:random_forest_n_heads_dist_2}
\end{figure*}

If we lower the threshold for separating ``near'' syntactic dependents from ``far'' dependents to $Dist > 2$, we observe the results in  
figures~\ref{fig:layer_perf_dist_2}  and~\ref{fig:random_forest_n_heads_dist_2} for ablation studies in Section~\ref{sec:ablation-studies}.
These figures are included here for comparison with Figures~\ref{fig:layer_perf_dist_3} and~\ref{fig:random_forest_n_heads_dist_3}, respectively.

\end{document}